\documentclass[journal]{IEEEtran}

\usepackage{cite}
\usepackage{graphicx}

\usepackage{soul}
\usepackage{ulem} %to strike the words
\usepackage{hyperref}

\usepackage{subfigure}
\usepackage{threeparttable}
\usepackage{caption}
\usepackage{setspace}

\usepackage{pifont}

\usepackage{orcidlink}

\usepackage{multirow}
\usepackage{pbox}
\usepackage{amsopn}
\usepackage{subcaption}
\usepackage{longtable}
\usepackage{array}
\usepackage{makecell}
\usepackage{tikz}
\usepackage{tabularray}
\usepackage{adjustbox} 
\usepackage{pifont}% 
\usepackage{colortbl}
\usepackage{caption}
\usepackage{capt-of}
\usepackage{booktabs}

\usepackage{color}
\definecolor{mypink1}{rgb}{0.858, 0.188, 0.478}
\definecolor{mygray}{gray}{0.6}
\definecolor{mygreen}{rgb}{0.53, 0.66, 0.42}
\definecolor{revisioncolor}{RGB}{0, 0, 0}  % Black
\newcommand{\revised}[1]{\textcolor{revisioncolor}{#1}}
\usepackage{amsfonts}
\usepackage{scalerel}
\usepackage{bbold}
\usepackage{amsmath}
\usepackage{mathtools}
\usepackage{amssymb}
\newcommand{\ie}{{\it i.e.}}
\newcommand{\eg}{{\it e.g.}}

\newcommand{\argmin}{\mathop{\rm arg~min}\limits}
\newcommand{\mathbbm}[1]{\textrm{\usefont{U}{bbm}{m}{n}#1}} 

\newcommand{\mycomment}[1]{%
}%

\usepackage{algorithm,algpseudocode}
\algnewcommand{\Inputs}[1]{%
\Statex \textbf{Inputs:}
  \Statex \hspace*{\algorithmicindent}\parbox[t]{.8\linewidth}{\raggedright #1}
}

\algnewcommand{\Outputs}[1]{%
\Statex\textbf{Outputs:}
  \Statex \hspace*{\algorithmicindent}\parbox[t]{.8\linewidth}{\raggedright #1}
}

\algnewcommand{\algorithmicforeach}{\textbf{for each}}
\algdef{SE}[FOR]{ForEach}{EndForEach}[1]
  {\algorithmicforeach\ #1\ \algorithmicdo}% \ForEach{#1}
  {\algorithmicend\ \algorithmicforeach}% \EndForEach

\begin{document}

\title{ \huge YOWO: You Only Walk Once to Jointly Map An Indoor Scene and Register Ceiling-mounted Cameras} 

\author{Fan Yang, Sosuke Yamao,  Ikuo Kusajima, Atsunori Moteki, Shoichi Masui, and Shan Jiang
        \thanks{
        Corresponding author: Fan Yang (fan.yang@fujitsu.com). }
        \thanks{
        All authors are with Fujitsu Research, Japan.} 
}

% and the end of the author line. \ywu{\emph{i.e.}}., if you had this:


%\markboth{IEEE Transactions on Circuits and Systems for Video Technology }%
%{F.Y. \MakeLowercase{\textit{et al.}}: YOWO}

\maketitle


\begin{abstract}
	Using ceiling-mounted cameras (CMCs) for indoor visual capturing opens up a wide range of applications. However, registering CMCs to the target scene layout presents a challenging task. While manual registration with specialized tools is inefficient and costly, automatic registration with visual localization may yield poor results when visual ambiguity exists. To alleviate these issues, we propose a novel solution for jointly mapping an indoor scene and registering CMCs to the scene layout. Our approach involves equipping a mobile agent with a head-mounted RGB-D camera to traverse the entire scene once and synchronize CMCs to capture this mobile agent. The egocentric videos generate world-coordinate agent trajectories and the scene layout, while the videos of CMCs provide pseudo-scale agent trajectories and CMC relative poses. By correlating all the trajectories with their corresponding timestamps, the CMC relative poses can be aligned to the world-coordinate scene layout. Based on this initialization, a factor graph is customized to enable the joint optimization of ego-camera poses, scene layout, and CMC poses. We also develop a new dataset, setting the first benchmark for collaborative scene mapping and CMC registration\footnote{\url{https://sites.google.com/view/yowo/home}.}
Experimental results indicate that our method not only effectively accomplishes two tasks within a unified framework, but also jointly enhances their performance. We thus provide a reliable tool to facilitate downstream position-aware applications. 
\end{abstract}

\begin{IEEEkeywords}
Multi-camera Video, Ego-centric Video, Indoor Scene Mapping, Camera Registration.
\end{IEEEkeywords}

% \fi
\IEEEpeerreviewmaketitle

\section{Introduction}\label{sec:intro}
  \IEEEPARstart{M}{ulti-camera} systems are commonly installed on ceilings in various indoor spaces, such as schools, retail stores, factories, and underground garages. Those scene-aware CMC captures have been employed for enhancing public safety surveillance, indoor virtual/augmented reality, and human-computer interaction~\cite{zhou2008activity, aghajan2009multi,multi-view_cap_2011, wang2013intelligent, xu2014efficient, mekhalfi2014compressive,3d_cap_ar_2015, li2017multi, dai2021indoor, sener2022assembly101, Naphade23AIC23}. To facilitate the functionality of scene-aware CMC capturing, a crucial aspect is mapping the scene layout and registering CMC six-degrees-of-freedom (6-DoF) poses to the scene layout.

  \begin{figure*}[th!]
	\centering
	\includegraphics[width=\textwidth]{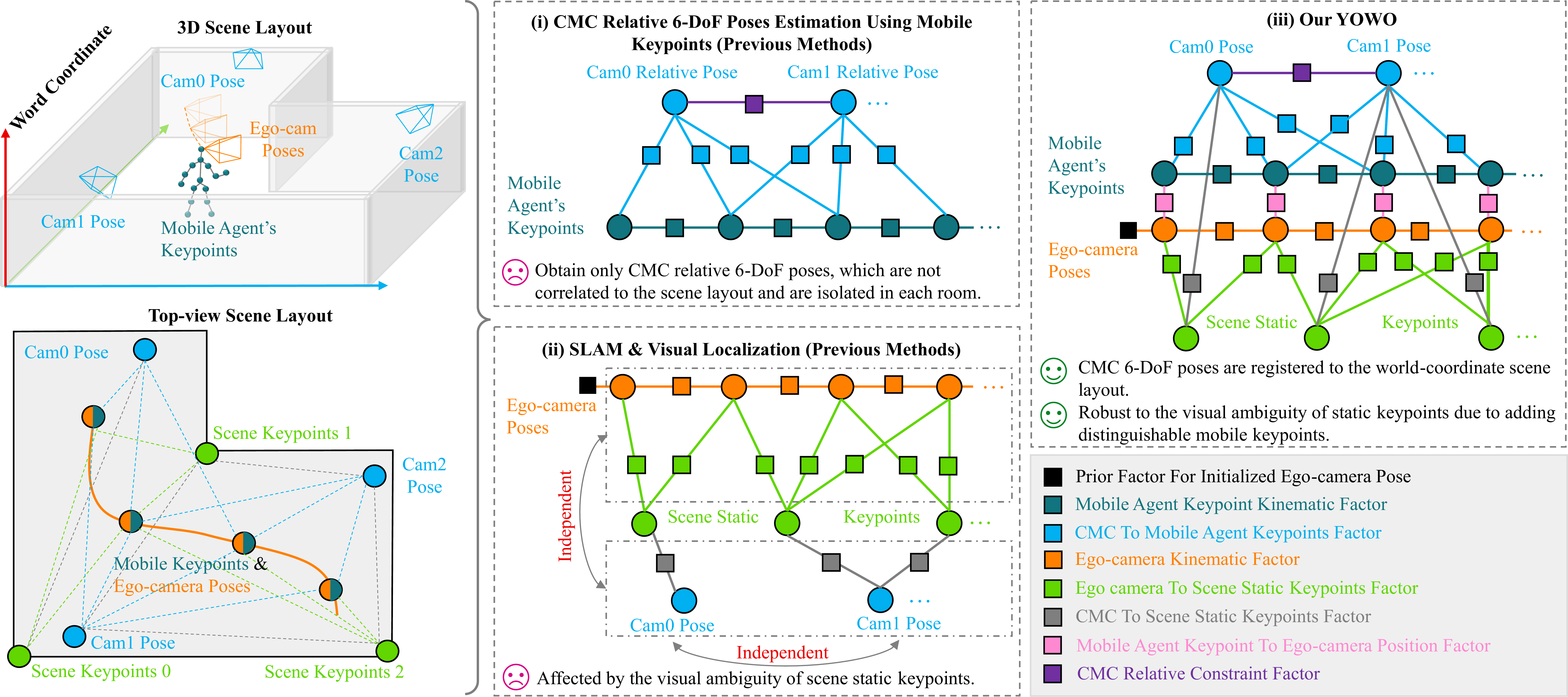}
	\caption{\textbf{Overview of scene mapping and CMC 6-DoF pose registration}. While using mobile keypoints~\cite{lee2022extrinsic, patzold2022online} can estimate robust CMC relative 6-DoF poses (i), the application of SLAM \& visual localization~\cite{taira2018inloc, sarlin2019coarse} can generate CMC 6-DoF poses registered to the scene (ii). We bring their merits into our YOWO (iii).}
	\label{fig:factor_graph}
  \end{figure*}  
  
  \begin{table*}[th!]
	\begin{center}
	  \caption{\textbf{Dataset comparison.} We offer a novel dataset for collaborative CMC and ego-camera capturing. }
	  \label{table:data_compare}
	  \rowcolors{2}{}{lightgray}
	  \begin{adjustbox}{width=\textwidth}
	  \begin{tabular}{llccccccc}
		\textbf{Dataset}   & \textbf{Target Task}    & \textbf{Scene Type}  & \begin{tabular}[c]{@{}l@{}} \textbf{No. Ego/Multi}\\ \textbf{Cameras}\end{tabular}   & \begin{tabular}[c]{@{}l@{}} \textbf{Multi-camera}\\ \textbf{Co-visibility}\end{tabular} & \textbf{Captures} & \begin{tabular}[c]{@{}l@{}} \textbf{Matching} \\ \textbf{Keypoints}\end{tabular} & \begin{tabular}[c]{@{}l@{}} \textbf{Estimated Camera} \\ \textbf{Vertical Height}\end{tabular}\\ \toprule
		OMECC~\cite{patzold2022online}      & \begin{tabular}[c]{@{}l@{}} Human Pose Estimation\\ \& CMC Relative  \\ 6-DoF Poses Estimation \end{tabular}      &Indoor    & 0, 20      &  \begin{tabular}[c]{@{}l@{}} Medium \\ \& High  \end{tabular}        &  Multi-view: RGB       & Mobile & \textgreater{} 2 m  \\ 
		InLoc~\cite{taira2018inloc}  & \begin{tabular}[c]{@{}l@{}}Scene Reconstruction \\ \& Visual Localization \end{tabular}  & Indoor  & 1, 0    &  -         & Ego-view: RGB-D   & Static &  \textless{} 2 m\\  
		Egobody~\cite{Zhang2022egobody}   & Human Pose Estimation  & Indoor  & 1, 5     &  High           & \begin{tabular}[c]{@{}l@{}} Multi-view: RGB-D \\ Ego-view: RGB-D  \end{tabular} & Static & \textless{} 2 m\\ 
	  Egohumans~\cite{khirodkar2023egohumans} & \begin{tabular}[c]{@{}l@{}}Human Pose Estimation\\ \& Multi-human Tracking\end{tabular}                & In/Outdoor  & 4, 15  &  High         & \begin{tabular}[c]{@{}l@{}} Multi-view: RGB \\ Ego-view: RGB  \end{tabular}  & Static & \textless{} 2 m\\ 
	  \textbf{Ours}      & \begin{tabular}[c]{@{}l@{}}Map Scene Layout \\ \& Register CMC \\ 6-DoF Poses To Scene\end{tabular}      &Indoor    & 1, 5-17     & \begin{tabular}[c]{@{}l@{}}Low  \\ \& Medium \\ \& High  \end{tabular}    & \begin{tabular}[c]{@{}l@{}} Multi-view: RGB \\ Ego-view: RGB-D  \end{tabular}         & \begin{tabular}[c]{@{}l@{}}Static \& \\  Mobile  \end{tabular} & \textgreater{} 2 m 
	\end{tabular}
  \end{adjustbox}
  \end{center}
  %\vspace{-0.4cm}
  \end{table*}
  
  In the field of scene mapping and CMC 6-DoF pose registration, two types of methodologies have been introduced. The first approach involves manual interventions: to map the scene layout using measuring tools (\eg, rulers), and, to register CMC 6-DoF poses using calibration devices (\eg, 3D calibration cubes) with established positions. Markers on the calibration devices are associated with CMCs to enable scene-aware 6-DoF pose registration~\cite{zhang2000flexible, schmidt2014calibration, munaro2016openptrack, rameau2022mc}. 
  However, this method is not scalable and proves to be inefficient.
  The second approach involves simultaneous localization and mapping (SLAM) and visual localization~\cite{davison2007monoslam, ataer2014calibration, kendall2015posenet, humenberger2022investigating, sarlin2022lamar, do2022learning, panek2022meshloc, wang2023ct, AR_loc_2024}. Following the 3D scene reconstruction via SLAM, it becomes feasible to align CMC 2D captures with the 3D scene features and subsequently derive correlated CMC 6-DoF poses (Fig.~\ref{fig:factor_graph} (ii)). However, this approach could potentially be vulnerable to visual ambiguity of scene features, especially when considering the disparity between CMC captures and ego-camera captures (Fig.~\ref{fig:visual_ambiguity}).

  In our pursuit of automating the process and mitigating visual ambiguity, we look into another category of works---CMC \textit{relative} 6-DoF pose estimation---that employs distinguishable mobile keypoints (\eg, human skeletons) as cross-camera matching features~\cite{puwein2015joint, xu2021wide, huang2021dynamic, charco2021camera, liu2022auto, patzold2022online, lee2022extrinsic, gordon2022flex, saini2023smartmocap, ci2023proactive, khirodkar2023egohumans}. They propose that mobile keypoints could sustain robustness despite variations in viewpoint and viewing conditions across all pertinent views. Even though those approaches exhibit a commendable capability to mitigate the visual ambiguity inherent in scene static keypoints, their resulting CMC 6-DoF poses have not been automatically aligned to the scene layout, and, CMCs in isolated rooms cannot be correlated (Fig.~\ref{fig:factor_graph} (i)).
  
  Drawing inspiration from the above pioneering research, we propose a novel framework to jointly optimize mapping an indoor scene and registering CMC 6-DoF poses to this scene. Specifically, our methodology involves equipping a mobile agent, such as a robot or a person, with a head-mounted RGB-D camera to traverse the entire scene. Meanwhile, CMCs are synchronized to capture this mobile agent. While the ego-camera captures yield agent trajectories and the scene layout in the world coordinate, the CMC captures provide pseudo-scale agent trajectories and CMC relative 6-DoF poses. By correlating all the trajectories with their corresponding timestamps, the CMC relative 6-DoF poses can be aligned to the world-coordinate scene layout. Based on this initialization, a factor graph~\cite{dellaert2012factor, dellaert2017factor} is tailored to enable the joint optimization of ego-camera 6-DoF poses, scene layout, and CMC 6-DoF poses (Fig.~\ref{fig:factor_graph} (iii)).
  Once the joint optimization is complete, we obtain a world-coordinate scene layout and CMC 6-DoF poses registered to this layout, which enables a wide range of subsequent applications. Our solution is user-friendly. As implied by our title, consider yourself as a mobile agent, then ``\textit{you only walk once (YOWO)}'' in the indoor scene to approach the results.

  As depicted in Tab.~\ref{table:data_compare}, our preferred settings markedly deviate from those of existing datasets, and therefore, we propose a novel dataset tailored to our specific requirements. We employ simulation engines, AI2THOR~\cite{kolve2017ai2, procthor} and Gym-UnrealCV~\cite{gymunrealcv2017, qiu2017unrealcv}, to generate synthetic data for collaborative CMC and ego-camera capture. Leveraging our custom dataset, we conducted a comparative analysis of our approach against well-established methodologies in the field. The experimental results underscore the effectiveness and efficiency of our YOWO.

  Our \textbf{contributions} are three-fold: 
  \begin{itemize}
	\item We propose a versatile solution for jointly mapping an indoor scene and registering CMCs to the scene layout. Existing studies have not jointly optimized these two tasks, which limits their performance and effectiveness.
	\item We contribute the first CMC and ego-camera collaborative capturing dataset, for the joint evaluation of scene layout mapping and CMC 6-DoF pose registration. 
	\item We perform a comprehensive evaluation of our method under diverse conditions, demonstrating that our approach can achieve robust and scalable performance. This capability facilitates a broad spectrum of downstream position-aware applications in indoor environments.
\end{itemize}

  %%%%%%%%%%%%%%%%%%%%%%%%%%%%%%%%%%%%%%%%%%%%%%%%%%%%%%%%%%%%%%%%%%%%%%%%%%%%%%%%
  
  \section{Related Work}

  \subsection{SLAM And Visual Localization}

  By employing the power of SLAM or Structure-from-Motion (SfM) methods~\cite{davison2007monoslam, snavely2008modeling, schonberger2016structure, Schops_2019_CVPR, campos2021orb, sun2021neuralrecon, zhu2022nice, wang2023ct}, the egocentric capture has been commonly applied for 3D indoor scene reconstruction~\cite{dai2017scannet, mao2022multiscan, zheng2022gimo, EgoLocate2023}. Once the indoor scene has been reconstructed, visual localization can be applied to estimate the most probable camera 6-DoF pose for an arbitrary capture~\cite{davison2007monoslam, ataer2014calibration, taira2018inloc, ding2019camnet, humenberger2020robust, brachmann2021visual, humenberger2022investigating, sarlin2022lamar, do2022learning, EgoLocate2023, AR_loc_2024}. However, static features extracted from the scene~\cite{detone2018superpoint,sun2021loftr}, which are relied upon by most visual localization methods, could be significantly affected by ambiguous textures, symmetry structures, or even the changing structures within an indoor scene~\cite{sattler2018benchmarking, taira2018inloc, spera2019egocart, wald2020beyond, humenberger2022investigating, sarlin2022lamar, arnold2022map,chen2022dfnet, matsumoto2024indoor}. 
  As shown in Fig.~\ref{fig:visual_ambiguity}, the disparity in perspective between the ego camera and CMCs further intensifies these challenges~\cite{humenberger2022investigating}. Furthermore, unlike the ego camera, our target CMCs remain static, and thus, we cannot leverage the global sequential camera poses to eliminate the camera localization errors~\cite{sarlin2022lamar}.
  Consequently, these issues undermine the performance of even the most accurate visual localization methods that follow a coarse-to-fine localization paradigm~\cite{sarlin2019coarse, sarlin2018leveraging, ding2019camnet, brachmann2019expert, humenberger2020robust, hausler2021patchnetvlad}.

  Besides, although observations from the ego camera and the CMCs could be strongly correlated under a synchronized configuration, extant literature has not yet leveraged this property. In our work, observations from the ego camera and CMCs are employed collaboratively to enhance both egocentric scene mapping and CMC registration. Specifically, instead of estimating independent CMC 6-DoF poses using conventional visual localization, observed mobile keypoints can be used together to form cross-view matching between CMCs and further infer their relative 6-DoF poses. Additionally, due to the lack of the global positioning signal, ego-camera SLAM may encounter position drift artifacts in indoor environments~\cite{davison2007monoslam, campos2021orb, EgoLocate2023}. Our method leverages CMC observations to generate pseudo-scale ego-camera trajectories, thereby reducing ego-camera SLAM drift errors through fusion.

  \subsection{CMC Relative Pose Estimation Using Mobile Keypoints} 
  
  Numerous previous studies~\cite{puwein2015joint, xu2021wide, huang2021dynamic, charco2021camera, su2021multi, ma2022virtual, liu2022auto, lee2022extrinsic, gordon2022flex, lee2022extrinsic, patzold2022online, saini2023smartmocap, ci2023proactive} have sought to automate the estimation of CMC relative pose by leveraging observed mobile keypoints. For instance, individuals observed across different cameras can be matched based on their appearance similarities, and the positions of these matched persons are used to estimate CMC relative 6-DoF poses~\cite{xu2021wide, su2021multi}. To improve the estimation performance, several investigations~\cite{huang2021dynamic, charco2021camera, ma2022virtual, ci2023proactive} utilize multi-person body keypoints to achieve more accurate results. To circumvent the challenges in cross-camera multi-person matching, another group of research~\cite{puwein2015joint, liu2022auto, lee2022extrinsic, patzold2022online, saini2023smartmocap} execute collaborative CMC relative 6-DoF pose estimation on a single person.
  
  Though these studies have significantly contributed to the advancement of CMC relative 6-DoF pose estimation, the capability to autonomously align CMC 6-DoF poses with the real scene layout has not been considered. In these works, scene mapping is treated as a separate task. However, this separation may limit potential applications. For example, in indoor surveillance, where map information is crucial, it is essential to register CMC 6-DoF poses to the scene layout. To remedy this limitation, our YOWO integrates the ego-camera scene mapping and CMC 6-DoF estimation into a unified framework to optimize them jointly.

  \subsection{Ego-camera And CMC Collaboration}

  \begin{figure}[th!]
	%\captionsetup{font=scriptsize}
	  \includegraphics[width=\linewidth]{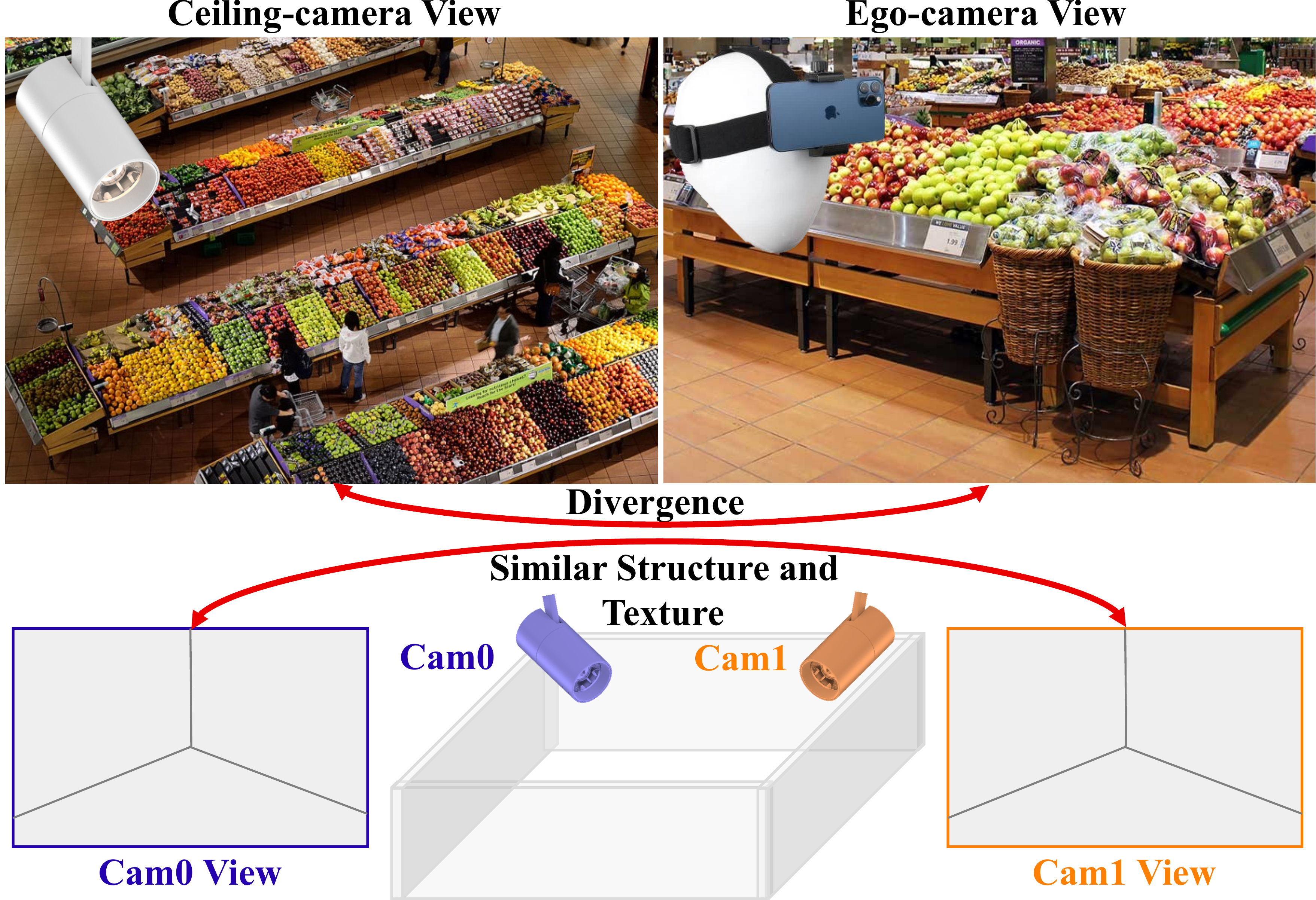}
	\caption{\textbf{Visual ambiguity.} Top: the divergence between CMC and ego-camera views in a supermarket scene. Bottom: the similar captures between different CMCs.}
	\label{fig:visual_ambiguity}
  \end{figure}
  
  Prior studies, as exemplified by~\cite{ardeshir2016ego2top, ardeshir2018egocentric, Zhang2022egobody, zheng2022gimo, araujo2023circle, bultmann2023external, khirodkar2023egohumans}, have made notable progress in exploring the collaboration between ego cameras and CMCs for human/robot pose estimation and tracking. Whereas our YOWO shares some similar hardware setup to these studies, we diverge from their focus and expand our scope to address a novel challenge: the collaboration between scene mapping and CMC 6-DoF poses registration. 
  
  In terms of the approach, previous studies obtain CMC 6-DoF poses as an independent step before proceeding ego-camera and CMC collaboration~\cite{ardeshir2018egocentric, Zhang2022egobody, araujo2023circle, bultmann2023external, khirodkar2023egohumans}. For instance, although Egohumans~\cite{khirodkar2023egohumans} targets human pose estimation with the collaboration of ego camera and CMCs, it applies a classical visual localization approach to obtain CMC 6-DoF poses before its human pose estimation. In contrast, our YOWO incorporates mobile keypoints to address the limitations of classical visual localization. Additionally, unlike existing methods that exclusively utilize factor graphs to optimize either scene mapping or CMC relative poses, our YOWO forms a novel factor graph to jointly optimize both.

  In terms of the dataset, several studies~\cite{Zhang2022egobody, khirodkar2023egohumans} do not fully capture scenes in their egocentric recording, which prevents the complete reconstruction of the scene layout. Some research~\cite{sturm2012benchmark, zheng2022gimo, araujo2023circle} opt for multi-camera setups to obtain ground-truth (GT) agent trajectories but exclude these multi-camera recordings from their released datasets. Some works, \eg,~\cite{li2020pose}, capture top-view videos utilizing mobile drone cameras other than our static CMCs. Other studies, \eg,~\cite{ardeshir2016ego2top}, capture only egocentric RGB videos, which is insufficient to recover the real-scale scene layout. Therefore, we specifically create a novel dataset for our experimental analysis, as shown in Tab.~\ref{table:data_compare}.

  %%%%%%%%%%%%%%%%%%%%%%%%%%%%%%%%%%%%%%%%%%%%%%%%%%%%%%%%%%%%%%%%%%%%%%%%%%%%%%%%

  \section{Approach}
  
  \begin{figure*}[t!]
	\centering
	\includegraphics[width=\textwidth]{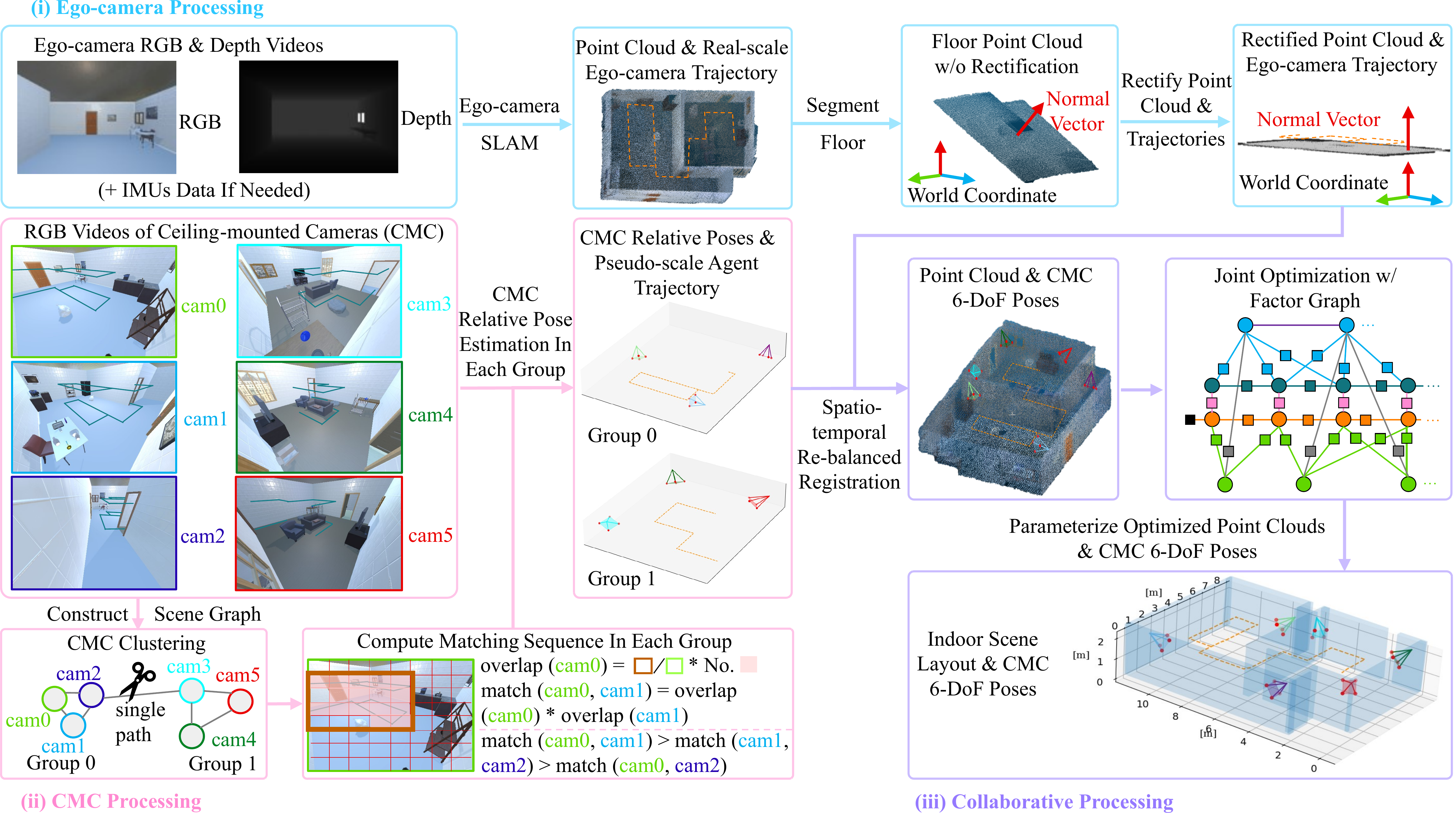}
	\vspace{-16pt}
	%\captionsetup{font=scriptsize}
	\caption{\textbf{Architecture of YOWO}. YOWO mainly includes three processes: ego-camera processing (Sec.~\ref{sec:ego_pro}), CMC processing (Sec.~\ref{sec:mul_pro}), and collaborative processing (Sec.~\ref{sec:col_pro}). RGB/RGB-D videos, possibly with Inertial Measurement Unit (IMU) data, are the inputs, while the outputs are the scene layout and CMC 6-DoF poses registered to it.}
  \label{fig:framework}
  \end{figure*}
  
  \noindent{\bf Problem formulation.}
  Our study includes an RGB-D ego camera with the index $c_e$ and multiple CMCs with indices $\left \{ c_0, \cdots, c_n \right \}$. 
  We suppose that all cameras have calibrated intrinsic parameters denoted as $\left \{ \mathbf{K}_{c_e}, \mathbf{K}_{c_0}, \cdots, \mathbf{K}_{c_n} \right \} \in \mathbb{R}^{3 \times 3}$, which are provided by camera logs, or, by using calibration algorithms~\cite{zhang2000flexible, bradski2000opencv}. 
  Our study yields two primary outputs: (i) The scene point clouds comprise $N_{\mathcal{M}}$ points, denoted by $\left\{ {}^{w}\mathbf{X}^{map}_{0},\cdots, {}^{w}\mathbf{X}^{map}_{N_{\mathcal{M}}}  \right\} \in \mathbb{R}^3$, which are further parameterized into 3D planes to represent the scene layout; (ii) The CMC 6-DoF poses registered to the scene layout, represented by matrices $\left \{\left[ {}^{w}\mathbf{R}_{c_0} |{}^{w}\mathbf{X}_{c_0}\right] , \cdots,  \left[ {}^{w}\mathbf{R}_{c_n} |{}^{w}\mathbf{X}_{c_n}\right] \right \} \in \mathbb{R}^{3 \times 4} $, where ${}^{w}\mathbf{R}_{c_i}$ and ${}^{w}\mathbf{X}_{c_i}$ are respectively the camera's orientation and position in the world coordinate. In this work, the superscripts $w$ and $l$ denote the coordinate systems for the world and the local, respectively.

  \noindent{\bf Overview.} YOWO consists of three primary processes: ego-camera processing, CMC processing, and collaborative processing (Fig.~\ref{fig:framework}). In Sec.~\ref{sec:ego_pro}, we introduce ego-camera processing, which includes using SLAM to acquire point clouds and ego-camera trajectories. We further elucidate the rectification process involved. We then detail CMC processing in Sec.~\ref{sec:mul_pro}, including CMC scene graph clustering and CMC relative pose estimation within each group. In Sec.~\ref{sec:col_pro}, we introduce a novel spatiotemporal rebalanced registration algorithm for correlating ego-camera and CMC trajectories, which facilitates the alignment of CMC poses with point clouds. Based on this initialization, a factor graph is customized to enable the collaborative optimization of ego-camera poses, scene layout, and CMC poses. The optimized results are transformed into parameterized indoor scene layout and CMC 6-DoF poses. 
  Given that our outputs are primarily designed for downstream applications, an offline approach is suitable.

  \subsection{Ego-camera Processing}
  \label{sec:ego_pro}

  We start by equipping a mobile agent with an ego camera positioned at the head, and let the agent traverse the entire scene to capture RGB-D images, denoted by $\left \{ \left \langle  \mathbf{I}^{rgb}_{c_e,f}, \mathbf{I}^{d}_{c_e,f} \right \rangle  \right \}_{\forall f \in \mathcal{F}}$, where $\mathcal{F}$ is the set of frames. Using $\left \{  \mathbf{I}^{d}_{c_e,f}\right \}_{\forall f \in \mathcal{F}} $, we initialize ego-camera motions using Iterative Closest Point (ICP)~\cite{chen1992object, rusinkiewicz2001efficient} with multiway registration~\cite{choi2015robust}. This approach circumvents the effect of visual noise and ambiguities in our initialization. Meanwhile, inertial measurement units (IMUs) attached to the ego camera could be employed to improve the results~\cite{campos2021orb, sarlin2022lamar}. 
By setting the initial world coordinate at the first ego-camera pose, the ego-camera motions are integrated into the ego-camera 6-DoF poses, as denoted by $\left \{ \left [{}^{w}\mathbf{R}_{c_e,f} \vert {}^{w}\mathbf{X}_{c_e,f} \right ] \right \}_{\forall f \in \mathcal{F}}$. Next, $\left \{  \mathbf{I}^{rgb}_{c_e,f}\right \}_{\forall f \in \mathcal{F}} $ are incorporated to refine the ego-camera 6-DoF poses. We apply keypoint detection and matching algorithms~\cite{detone2018superpoint, sarlin2020superglue, sun2021loftr} to obtain matched 2D keypoints from consecutive frames. To address challenging indoor appearance, in addition to directly detecting 2D scene keypoints, we also apply a line detection and matching method~\cite{pautrat2021sold2} to obtain matched line endpoints. To this end, 2D scene keypoints, captured by the ego camera, are represented by $\left\{ \mathbf{x}_{c_e,f,m} \right \}_{\forall m \in \mathcal{M},\forall f \in \mathcal{F}}$, where $\mathcal{M}$ denotes the set of scene keypoint indices. In addition to using the Epipolar constraint~\cite{xu1996epipolar}, the initialized ego-camera poses and depth information are employed to filter out inconsistent keypoints~\cite{sun2022improving}. Specifically, based on preset ego-camera poses, we separately estimate 3D points by using paired 2D keypoint triangulation and depth-to-3D mapping~\cite{hartley2003multiple}. If the Euclidean distance between identical 3D points exceeds a certain threshold, we reject the corresponding 2D keypoints~\cite{hartley2003multiple}. Filtered 2D keypoints and corresponding 3D points form 2D-3D matches $\left \{ \left \langle \mathbf{x}_{c_e,f,m}, {}^{w}\mathbf{X}^{map}_{m} \right \rangle \right \}_{\forall m \in \mathcal{M},\forall f \in \mathcal{F}}$. 
\revised{For the sake of simplicity, we employ a basic method to initialize scene reconstruction in YOWO. However, it is flexible to adopt more advanced SfM/SLAM techniques to construct $\left \{ \left \langle \mathbf{x}_{c_e,f,m}, {}^{w}\mathbf{X}^{map}_{m} \right \rangle \right \}_{\forall m \in \mathcal{M},\forall f \in \mathcal{F}}$ and  $\left \{ \left [{}^{w}\mathbf{R}_{c_e,f} \vert {}^{w}\mathbf{X}_{c_e,f} \right ] \right \}_{\forall f \in \mathcal{F}}$.}

  We apply a bundle adjustment (BA)~\cite{triggs2000bundle, hartley2003multiple} 
  to jointly adjust ego-camera poses and 2D-3D matches, which is formulated by
  %\begin{scriptsize}
\begin{equation}\label{eq:ego_bd}
	\begin{split}
	 \argmin_{ \substack{ \left \{ \left [{}^{w}\mathbf{R}_{c_e,f} \middle | {}^{w}\mathbf{X}_{c_e,f} \right ] \right \},\\ \left \{ {}^{w}\mathbf{X}^{map}_{m} \right \}}} &
	 \sum_{\forall f \in \mathcal{F}}  \sum_{\forall m \in \mathcal{M} }   \left\| \varepsilon_{ego} \left (\left [{}^{w}\mathbf{R}_{c_e,f} \middle | {}^{w}\mathbf{X}_{c_e,f} \right ], \right.\right.\\
	 &\left.\left.{}^{w}\mathbf{X}^{map}_{m} \right ) \right\|^2, 
	\end{split}
\end{equation}
where the error to be minimized are represented by
\begin{equation}\label{eq:ego_bd_error}
	\begin{split}
	 &\left\|\varepsilon_{ego} \left ( \left [{}^{w}\mathbf{R}_{c_e,f} \middle | {}^{w}\mathbf{X}_{c_e,f} \right ], {}^{w}\mathbf{X}^{map}_{m} \right ) \right\|^2 \\
	 =& v_{f,m}~  \rho  \left\| \mathbf{x}_{c_e,f,m} 
	 - \pi \left (\mathbf{K}_{c_e}\left [{}^{w}\mathbf{R}_{c_e,f}^{\top}  \middle | -{}^{w}\mathbf{R}_{c_e,f}^{\top}  {}^{w}\mathbf{X}_{c_e,f} \right ], \right.\right.  \\
	 & \left. \left.
	 {}^{w}\mathbf{X}^{map}_{m}\right ) \right\|^2,\\
	\end{split}
\end{equation}
where $\rho$ is a Huber robust M-estimator, $\pi(\cdot)$ represents the projection function that maps 3D points to 2D points given the camera pose. The binary variable $v_{f,m}$ equals to $1$ if ${}^{w}\mathbf{X}^{map}_{m}$ is visible at  frame $f$, and $0$ otherwise. The formula is optimized by employing the Levenberg–Marquardt algorithm~\cite{more2006levenberg}. We construct the point clouds using the optimized ego-camera poses and RGB-D images.
  
In practice, it is challenging to perfectly initialize the ego-camera pose such that its vertical direction is perpendicular to the ground plane. The alignment bias is subsequently inherited by the constructed point clouds (Fig.~\ref{fig:framework}). To accurately map the scene layout, we compute a rotation matrix $\mathbf{R}_{rect}$ to rectify the point clouds in the vertical direction. We apply planar patch detection~\cite{AraujoOliveira_2020a} to detect planes on point clouds. The plane with the most points is considered as the ground plane. Let $\mathbf{v}_{gp}$ represent the normal vector of the ground plane, $\mathbf{v}_{w}$ denote the initialized world-coordinate vertical vector, and $\mathbb{I}_{3 \times 3}$ symbolize an identity matrix, the rotation matrix is given by
%\begin{scriptsize}
  \begin{equation}\label{eq:rot_gp2w}
  \begin{split}
    \mathbf{R}_{rect} &= \mathbb{I}_{3 \times 3} + [\mathbf{g}]_{\times} + [\mathbf{g}]_{\times}^2\frac{1-\mathbf{v}_{gp} \cdot \mathbf{v}_{w}}{\left \| \mathbf{g} \right \|^2}, \\
	\textrm{with}~  \mathbf{g} &= \mathbf{v}_{gp} \times \mathbf{v}_{w}~\textrm{and}
    ~ [\mathbf{g}]_{\times} \stackrel{\rm def}{=} \begin{bmatrix}
      \,\,0 & \!-\mathbf{g}_3 & \,\,\,\mathbf{g}_2\\
      \,\,\,\mathbf{g}_3 & 0 & \!-\mathbf{g}_1\\
      \!-\mathbf{g}_2 & \,\,\mathbf{g}_1 &\,\,0
      \end{bmatrix}.
  \end{split}
  \end{equation}

  We then use $\mathbf{R}_{rect}$ to rectify $\left \{ {}^{w}\mathbf{X}^{map}_{m} \right \}$ and $\left \{ \left [{}^{w}\mathbf{R}_{c_e,f} \middle | {}^{w}\mathbf{X}_{c_e,f} \right ] \right \}$ in the vertical direction:
  %\begin{scriptsize}
	\begin{align}
	  {}^{w}\mathbf{X}^{map}_{m} = & \mathbf{R}_{rect} {}^{w}\mathbf{X}^{map}_{m} \label{eq:rot_1} \\ 
	  \left[ {}^{w}\mathbf{R}_{c_e,f} \middle |  {}^{w}\mathbf{X}_{c_e,f}   \right] = & \mathbf{R}_{rect} \left[  {}^{w}\mathbf{R}_{c_e,f} \middle |  {}^{w}\mathbf{X}_{c_e,f}   \right]\label{eq:rot_2}.
	\end{align}  
  %\end{scriptsize}

	\subsection{CMC Processing}
  \label{sec:mul_pro}
  Our mobile agent can either be a human or a robot. To estimate the 2D keypoint sequence of the agent for each CMC, we employ off-the-shelf 2D keypoint estimation and tracking algorithms~\cite{xiao2018simple, sun2019deep, zhang2022bytetrack, ultralytics_yolov8} for a human agent. For a robot agent, we initially annotate samples for its top and bottom points, subsequently fine-tuning the same algorithms to estimate its keypoints. We skip the details of annotation and training here. To eliminate jitters in the mobile keypoint sequence, we further apply a Savitzky–Golay filter~\cite{savitzky1964smoothing} to smooth the data.

  Given our assumption that there is only one mobile agent in the scene, we can establish cross-camera correspondences by referencing the timestamps and keypoint indices. For instance, at the frame $f$, if a mobile keypoint with index $o$ is concurrently observed by a set of ceiling-mounted cameras $\mathcal{C}^{cmc}$, the 2D keypoint set $\left\{\mathbf{x}_{c_i, f,o} \right\}_{\forall c_i \in \mathcal{C}^{cmc} }$ corresponds to the same 3D keypoint $\mathbf{X}^{cmc}_{a,f,o}$. Thus, although we follow incremental SfM~\cite{snavely2008modeling, schonberger2016structure} to estimate CMC relative 6-DoF poses, we can skip the steps of image retrieval~\cite{arandjelovic2016netvlad, gordo2017end} and feature matching~\cite{muja2009fast, sarlin2020superglue, sun2021loftr}. 
  However, due to the physical barriers posed by indoor walls, CMCs may be divided into separate spaces, and therefore, we cluster them into groups based on their co-observations~\cite{strasdat2011double, sarlin2018leveraging, sarlin2019coarse}. Given the unique nature of mobile keypoints, we let $\mathcal{A}$ and $\mathcal{B}$ respectively denote the sets of co-visible mobile keypoints of camera $c_i$ and $c_j$, as determined by their timestamps and keypoint indices. We then exclude noisy samples by estimating the fundamental matrix ${}^{c_j}\mathbf{F}_{c_i}$ and verifying the Epipolar constraint error with a MAGSAC scheme~\cite{barath2020magsac++} (\ie, a threshold-unsensitive RANSAC~\cite{fischler1981random}).
  We formulate the inlier counting for $\mathcal{A}$ and $\mathcal{B}$ as
  %\begin{scriptsize}
	\begin{equation}\label{eq:inliers}
		\begin{split}
	  & \textnormal{inliers} (c_i, c_j) = \underset{\forall \mathbf{x}_{c_i,f,o} \in \mathcal{A}~~ \forall \mathbf{x}_{c_j,f,o} \in \mathcal{B}}{\sum}  \\ &\mathbbm{1}  \left [ \left \|\mathbf{x}_{c_i,f,o}^{\top} {}^{c_j}\mathbf{F}_{c_i} \mathbf{x}_{c_j,f,o} \right \|_2 \leq \textnormal{RANSAC Threshold} \right ].
		\end{split}
  \end{equation}
  %\end{scriptsize}
  Using Eq.~\ref{eq:inliers}, we construct a scene graph~\cite{kushal2012visibility, schonberger2016structure} and perform graph clustering~\cite{schaeffer2007graph} to divide CMCs into groups. Within the same group, we impose two conditions between any two elements: (i) the presence of at least eight matching pairs~\cite{hartley1997defense, hartley2003multiple}, (ii) a minimum of two paths connecting one element to another, to ensure the 2D-3D matches are available for any camera. We apply these criteria to guarantee that stable CMC relative poses can be formed.

  Within a camera group, we assume that the inlier samples of $\mathcal{A}$ and $\mathcal{B}$ form new sets $\tilde{\mathcal{A}}$ and $\tilde{\mathcal{B}}$. Inspired by~\cite{irschara2009structure, schonberger2016structure, rau2020predicting, sarlin2022lamar}, we define a binary grid matrix $\mathbf{G}_{c_i}$ to quantize $\tilde{\mathcal{A}}$ and formulate our spatial overlap score as
  %\begin{scriptsize}
	\begin{align}
	  \mathbf{G}_{c_i}[r, l] = \begin{cases}
		1, & \textrm{if } \exists~ \mathbf{x}_{c_i,f,o} \in [r, r+1) \times [l, l+1), \\
		 & \textrm{and}~\forall \mathbf{x}_{c_i,f,o} \in \tilde{\mathcal{A}}, \\
		0, & \textrm{otherwise}
		\end{cases} \label{eq:grid} 
	\end{align}
	\begin{align}\nonumber
	  \textnormal{overlap} (c_i) &=  \frac{\left (\underset{r}{\textrm{max}}\mathbf{G}_{c_i}- \underset{r}{\textrm{min}}\mathbf{G}_{c_i} \right)  \cdot \left (\underset{l}{\textrm{max}}\mathbf{G}_{c_i} - \underset{l}{\textrm{min}} \mathbf{G}_{c_i} \right)} {N_{GW} \cdot N_{GH}} \\  
	  &\cdot \sum_{r \in N_{GH}, l \in N_{GW}} \mathbf{G}_{c_i}[r, l], \label{eq:overlap}
	\end{align}
  %\end{scriptsize}
  where $r \in N_{GH}$ and $l \in N_{GW}$ denote the row and column of $\mathbf{G}_{c_i}$, respectively, and $[r, r+1) \times [l, l+1)$ represents the grid cell. A higher overlap score generally results in a more accurate 6-DoF pose estimation. 
  We obtain the $\textnormal{overlap} (c_j)$ for $\tilde{\mathcal{B}}$ with the same formula. 
  
  Since CMCs are typically installed with well-structured triangulation baselines and angles, we leave these factors aside. Thus, we select the initial pair of cameras with the largest matching score, computed using 
  %\begin{scriptsize}
	\begin{equation}\label{eq:match}
	\begin{split}
	  \textnormal{match} (c_i,c_j) =  \textnormal{overlap} (c_i) \cdot \textnormal{overlap} (c_j).
	\end{split}
	\end{equation}
  %\end{scriptsize}
  Assuming the camera indices of the initial pair are $c_i$ and $c_j$, we compute their essential matrix ${}^{c_j}\mathbf{E}_{c_i}$ from previously obtained ${}^{c_j}\mathbf{F}_{c_i}$, as 
  %\begin{scriptsize}
	\begin{equation}\label{eq:F2E}
	\begin{split}
  {}^{c_j}\mathbf{E}_{c_i}=\mathbf{K}_{c_i}^{\top}{}^{c_j}\mathbf{F}_{c_i}\mathbf{K}_{c_j}.
	\end{split}
	\end{equation}
  %\end{scriptsize}
  We then extract the rotation matrix ${}^{c_j}\mathbf{R}_{c_i}$ and the translation vector ${}^{c_j}\mathbf{t}_{c_i}$ by applying the singular value decomposition (SVD) to ${}^{c_j}\mathbf{E}_{c_i}$~\cite{xu1996epipolar, hartley2003multiple}. Among multiple solutions, we employ geometric verification to select the most suitable one~\cite{park2017colored}. The center of the reference camera $c_i$ is selected as the local coordinate center for its group, as ${}^{l}\mathbf{R}_{c_i}=[\mathbb{I}_{3 \times 3}] $ and ${}^{l}\mathbf{X}_{c_i}=[0,0,0]$. The relative 6-DoF of camera $c_j$ is $[{}^{l}\mathbf{R}_{c_j} \vert {}^{l}\mathbf{X}_{c_j}]$, with ${}^{l}\mathbf{R}_{c_j} = {}^{c_j}\mathbf{R}_{c_i}^{\top} $ and ${}^{l}\mathbf{X}_{c_j}=-{}^{c_j}\mathbf{R}_{c_i}^{\top}{}^{c_j}\mathbf{t}_{c_i}$. We further construct their projection matrices, as
  %\begin{scriptsize}
	\begin{align}\label{eq:proj}
	  \mathbf{P}_{c_i} = \mathbf{K}_{c_i} [{}^{l}\mathbf{R}_{c_i}^{\top} \vert -{}^{l}\mathbf{R}_{c_i}^{\top}{}^{l}\mathbf{X}_{c_i}],\\
	  \mathbf{P}_{c_j} = \mathbf{K}_{c_j}[{}^{l}\mathbf{R}_{c_j}^{\top} \vert -{}^{l}\mathbf{R}_{c_j}^{\top}{}^{l}\mathbf{X}_{c_j}],
	\end{align}
  %\end{scriptsize}
  %$\mathbf{P}_{c_i} = \mathbf{K}_{c_i} [{}^{l}\mathbf{R}_{c_i}^{\top} \vert -{}^{l}\mathbf{R}_{c_i}^{\top}{}^{l}\mathbf{X}_{c_i}]$ and $\mathbf{P}_{c_j} = \mathbf{K}_{c_j}[{}^{l}\mathbf{R}_{c_j}^{\top} \vert -{}^{l}\mathbf{R}_{c_j}^{\top}{}^{l}\mathbf{X}_{c_j}]$, 
  and apply 3D triangulation~\cite{hartley2003multiple, schonberger2016structure} to generate the local-coordinate 3D mobile keypoints $\left\{{}^{l}\mathbf{X}^{cmc}_{a,f,o} \vert \forall f \in \mathcal{F},\forall o \in \mathcal{O} \right\}$, where $\mathcal{O}$ is the set of mobile keypoint indices.
  
  Leveraging the timestamps and keypoint indices, we can establish 2D-3D matches between the estimated 3D points and the 2D points from a newly introduced camera within the same group. These matches can then be utilized to estimate the relative 6-DoF pose of the new camera, via P$n$P~\cite{kneip2011novel, zheng2013revisiting} and RANSAC. Among multiple unmatched cameras, we prefer to initially match the camera that has the largest overlapping score between its 2D points to the existing 3D points. As such, we utilize Eqs.~\ref{eq:grid} and \ref{eq:overlap} to compute overlapping scores for all unmatched cameras within a group, and select the camera with the highest overlapping score to estimate its 6-DoF pose. Upon obtaining the new 6-DoF pose, we generate additional 3D mobile keypoints by integrating the new camera into the system. The overlapping scores should be updated after adding new 3D points. Repeating this process yields the full CMC relative 6-DoF poses and pseudo-scale agent trajectories within each group.

  After obtaining CMC relative 6-DoF poses, we perform a BA to refine them within each group. In our CMC BA, we denote the  error function for the 2D-3D projection constraint as $\varepsilon_{cmc}(\cdot)$.  We minimize it to jointly optimize CMC relative 6-DoF poses $\left \{ \left [{}^{l}\mathbf{R}_{c_i} \vert {}^{l}\mathbf{X}_{c_i} \right ] \right \}$ and 3D mobile keypoints $\left\{{}^{l}\mathbf{X}^{cmc}_{a,f,o}\right\}$ in their pseudo-scale local coordinates. The formulation is as follows:
  %\begin{scriptsize}
	\begin{equation}\label{eq:cmc_bd}
	\begin{split}
	  & \argmin_{ \left \{ \left [{}^{l}\mathbf{R}_{c_i} \middle | {}^{l}\mathbf{X}_{c_i} \right ] \right \}, \left \{ {}^{l}\mathbf{X}^{cmc}_{a,f,o} \right \}} \sum_{\forall c_i \in \mathcal{C}^{cmc}} \sum_{\forall f \in \mathcal{F}} \sum_{\forall o \in \mathcal{O}} 
	      \left\|  \varepsilon_{cmc}\left (\left [{}^{l}\mathbf{R}_{c_i} \middle | {}^{l}\mathbf{X}_{c_i} \right ], \right.\right.\\
	   & ~~~~~~~~~~~~~~~~~~~~~~~~~~~~~~~~~~~~~~~~~~~~~~~~~\left.\left.  
	   {}^{l}\mathbf{X}^{cmc}_{a,f,o} \right ) \right\|^2,\\
	 & \textrm{where}\\
	 & \left\| \varepsilon_{cmc} \left ( \left [{}^{l}\mathbf{R}_{c_i} \middle | {}^{l}\mathbf{X}_{c_i} \right ], {}^{l}\mathbf{X}^{cmc}_{a,f,o} \right ) \right\|^2  \\
	 & =  v_{c_i,f,o}~ \rho \left\| \mathbf{x}_{c_i,f,o} - \pi \left (\mathbf{K}_{c_i}\left [{}^{l}\mathbf{R}_{c_i}^{\top} \middle | -{}^{l}\mathbf{R}_{c_i}^{\top} {}^{l}\mathbf{X}_{c_i} \right ], 
	  {}^{l}\mathbf{X}^{cmc}_{a,f,o}\right ) \right\|^2.
	\end{split}
	\end{equation}
  %\end{scriptsize}
  Within a group, the set of cameras, frames, and keypoints are denoted by $\mathcal{C}^{cmc}$, $\mathcal{F}$, and $\mathcal{O}$. The binary variable $v_{c_i,f,o}$ represents the visibility of 3D-2D projection.  
  %While the convectional SfM may need to process thousands of 6-DoF poses, our target task generally involves a limited number of CMC. Therefore, the BA complexity in our method will not grow unbounded.

  \subsection{Collaborative Processing}
  \label{sec:col_pro}
  
  \noindent{\bf Register CMCs to the scene}. 
  Using multiple mobile keypoints provides sufficient observations to apply geometric verifications for a robust CMC relative 6-DoF pose estimation. However, in our ego-camera processing, we only can obtain the ego-camera trajectory $\left \{ {}^{w}\mathbf{X}_{c_e,f} \in \mathbb{R}^{3} \right \}$ for one position, which is closed to the agent's top keypoint in our setting. Therefore, we employ the agent's top keypoint trajectory, as denoted by $\left \{ {}^{l}\mathbf{X}^{cmc}_{a,f,o_{t}} \in \mathbb{R}^{3} \right \}$, to approximate the local-coordinate ego-camera trajectory. We then align $\left \{ {}^{l}\mathbf{X}^{cmc}_{a,f,o_{t}}  \right \}$ with $\left \{ {}^{w}\mathbf{X}_{c_e,f} \right \}$ to transform the CMC processing results to the world coordinate. Although there may exist a tiny discrepancy between the ego-camera position and the agent's top keypoint, this discrepancy becomes insignificant when CMCs are sufficiently distant from the agent. Furthermore, we will incorporate this discrepancy into our collaborative factor graph optimization with an appropriate uncertainty to tolerate it (see Eqs.~\ref{eq:f_e2e}~and~\ref{eq:factor_graph_all}).
  
  In previous studies~\cite{zhan2020visual, patzold2022online}, the Kabsch-Umeyama (KU) algorithm~\cite{kabsch1976solution, umeyama1991least} has been utilized to align relative CMC positions to the GT for evaluation purposes, or to realign relative CMC positions for iterative pose refinement. In our YOWO, the KU algorithm could be used for matching $\left \{ {}^{l}\mathbf{X}^{cmc}_{a,f,o_{t}} \right \}$ to $\left \{ {}^{w}\mathbf{X}_{c_e,f}  \right \}$.
  We want to find the optimal rescaling factor ${}^{l}s_{w}$, rotation matrix ${}^{l}\mathbf{R}_{w}$, and translation vector ${}^{l}\mathbf{t}_{w}$ such that:
  %\begin{scriptsize}
  \begin{equation}
	\label{eq:KU_origianl}
	\begin{split}
	  \argmin_{{}^{l}s_{w},{}^{l}\mathbf{R}_{w},{}^{l}\mathbf{t}_{w}} \sum_{\forall f \in \mathcal{F}}  \left \| {}^{w}\mathbf{X}_{c_e,f} - \left({}^{l}s_{w} {}^{l}\mathbf{R}_{w} {}^{l}\mathbf{X}^{cmc}_{a,f,o_{t}} + {}^{l}\mathbf{t}_{w}\right) \right \|^2.
  \end{split}  
  \end{equation}
  %\end{scriptsize}

\begin{figure}[tb!]
	%\captionsetup{font=scriptsize}
	\begin{center}
	  \includegraphics[width=\linewidth]{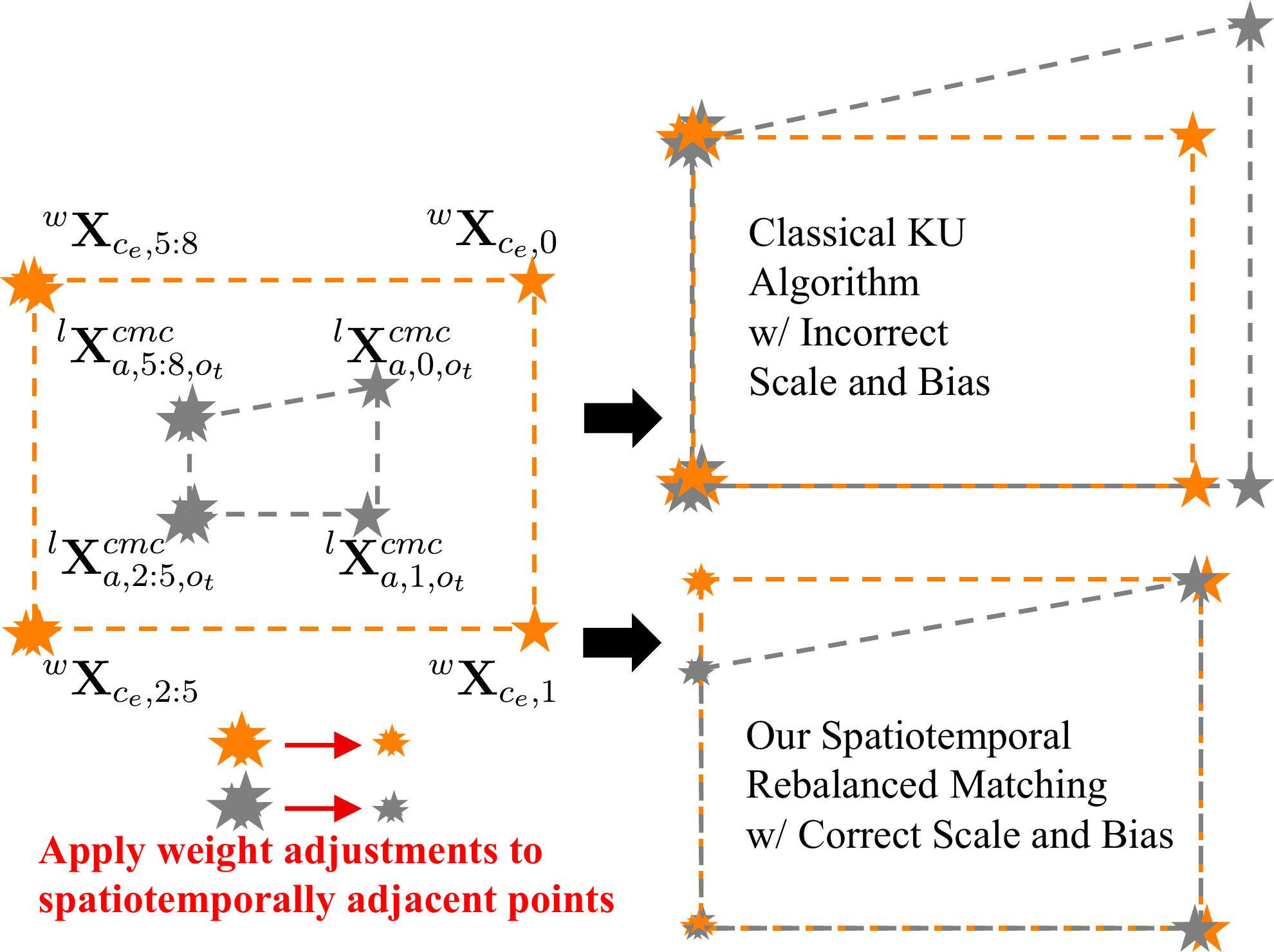}
	\end{center}
	\caption{\textbf{Comparison of classical KU matching and our spatiotemporal rebalanced matching.}}
	\label{fig:rebalanced_match}
  \end{figure}

  However, using Eq.~\ref{eq:KU_origianl} in YOWO presents a unique challenge: if the agent remains stationary for an extended period while capturing different views, the reference to timestamps will yield a large number of 3D points at the same position. This results in an overemphasis on this position, leading to biased matching, as depicted in Fig.~\ref{fig:rebalanced_match}. 

To mitigate this issue, we recast the KU algorithm to a spatiotemporal rebalanced matching algorithm. First, we adapt the single-linkage clustering~\cite{murtagh2012algorithms} to allocate $\left \{ {}^{w}\mathbf{X}_{c_e,f}  \right \}$ into clusters $\left \{ \phi_0, \phi_1, \cdots, \phi_{N_{clus}-1}  \right \}$. This adaptation incorporates two conditions into the single-linkage clustering process, ensuring that any two arbitrary elements, $\left \{ {}^{w}\mathbf{X}_{c_e,f_i},  {}^{w}\mathbf{X}_{c_e,f_j}  \right \} \in \phi_k$, fulfill the following criteria:
  %\begin{scriptsize}
  \begin{equation}
	\label{eq:clustering}
	\begin{split}
	  \begin{cases} 
		\left \|  {}^{w}\mathbf{X}_{c_e,f_i}-{}^{w}\mathbf{X}_{c_e,f_j} \right \|_2 \leq \textnormal{spatialThreshold} \\
		\left |f_i - f_j  \right | \leq \textnormal{temporalThreshold} 
	  \end{cases}
  \end{split}  
  \end{equation}
  %\end{scriptsize}
  Then, by using the clustering results, we add the rebalanced weight $\alpha_{f}$ to Eq.~\ref{eq:KU_origianl} and construct our rebalanced matching algorithm as
  %\begin{scriptsize}
  \begin{equation}
	\label{eq:reblanced_matching}
	\begin{split}
	  %{}^{l}_{w}\hat{s},{}^{l}_{w}\hat{\mathbf{R}},{}^{l}_{w}\hat{\mathbf{t}}  =& 
	 \argmin_{{}^{l}s_{w},{}^{l}\mathbf{R}_{w},{}^{l}\mathbf{t}_{w}} & \sum_{\forall f \in \mathcal{F}}  {\color{red} \alpha_{f}}  \left \| {}^{w}\mathbf{X}_{c_e,f} - ({}^{l}s_{w} {}^{l}\mathbf{R}_{w} {}^{l}\mathbf{X}^{cmc}_{a,f,o_{t}} + {}^{l}\mathbf{t}_{w}) \right \|^2,\\
	\textrm{where}~{\color{red} \alpha_{f}} & = \frac{1}{N_{clus}N_{\phi_k}}~\textrm{if}~{}^{w}\mathbf{X}_{c_e,f} \in \phi_k,\\
	 \sum_{\forall f \in \mathcal{F}} {\color{red} \alpha_{f}} &=1, 
  \end{split}  
  \end{equation}
  %\end{scriptsize}
  where $\mathcal{F}$ denotes the co-visible frames between the ego camera and the CMCs, $N_{clus}$ and $N_{\phi_k}$ represent the number of spatiotemporal clusters and elements of cluster $\phi_k$, respectively.

\revised{The corresponding solution slightly deviates from the original KU algorithm and can be derived using the following steps. We begin by calculating the weighted centroids with $\left \{  \alpha_{f} \right \}_{\forall f \in \mathcal{F}}$:
  \begin{equation}
	\begin{split}
	  {}^{l}\mathbf{\bar{X}}^{cmc}_{a,o_{t}}  = \sum_{\forall f \in \mathcal{F}} \alpha_{f} {}^{l}\mathbf{X}^{cmc}_{a,f,o_{t}} , ~
	{}^{w}\mathbf{\bar{X}}_{c_e} = \sum_{\forall f \in \mathcal{F}}  \alpha_{f} {}^{w}\mathbf{X}_{c_e,f}.
  \end{split}  
  \end{equation}
  Next, we compute the weighted cross-covariance matrix between matching sequences $\left \{{}^{w}\mathbf{X}_{c_e,f} \right \}_{\forall f \in \mathcal{F}} $ and $\left \{{}^{l}\mathbf{X}^{cmc}_{a,f,o_{t}} \right \}_{\forall f \in \mathcal{F}} $:
  \begin{equation}
	%\small
	\begin{split}
	\mathbf{H} =& 
	\begin{bmatrix}
	  {}^{w}\mathbf{X}_{c_e,0}  - {}^{w}\mathbf{\bar{X}}_{c_e}\\ 
	  {}^{w}\mathbf{X}_{c_e,1}  - {}^{w}\mathbf{\bar{X}}_{c_e}\\
	  \vdots\\
  {}^{w}\mathbf{X}_{c_e,N_{\mathcal{F}}-1}  - {}^{w}\mathbf{\bar{X}}_{c_e}
	  \end{bmatrix}^\intercal
					  \text{diag}\left (\alpha_0,\alpha_1,\cdots, \alpha_{N_{\mathcal{F}}-1} \right ) \\
	 & \begin{bmatrix}
						{}^{l}\mathbf{X}^{cmc}_{a,0,o_{t}}  - {}^{l}\mathbf{\bar{X}}^{cmc}_{a,o_{t}}\\ 
						{}^{l}\mathbf{X}^{cmc}_{a,1,o_{t}}  - {}^{l}\mathbf{\bar{X}}^{cmc}_{a,o_{t}}\\
										\vdots\\
						{}^{l}\mathbf{X}^{cmc}_{a,N_{\mathcal{F}}-1,o_{t}}  - {}^{l}\mathbf{\bar{X}}^{cmc}_{a,o_{t}}
										\end{bmatrix}                  
  ,
  \end{split}  
  \end{equation}
where $N_{\mathcal{F}}$ represent the number of the set $\mathcal{F}$. We then compute the weighted covariance of sequences $\left \{{}^{w}\mathbf{X}_{c_e,f} \right \}_{\forall f \in \mathcal{F}} $:
  \begin{equation}
	\begin{split}
	  \sigma_{{}^{w}\mathbf{X}_{c_e}}^2 = \sum_{\forall f \in \mathcal{F}} \alpha_f \left \|{}^{w}\mathbf{X}_{c_e,f}  - {}^{w}\mathbf{\bar{X}}_{c_e}\right \|^2 
  \end{split}  
  \end{equation}
   The remaining calculations are identical to the original Kabsch-Umeyama algorithm:
  \begin{equation}
	%\small
	\begin{split}
	\mathbf{H} =\mathbf{U} \Sigma \mathbf{V}^\intercal ,~
	d = \text{sign}\big(\text{det}\left(\mathbf{V}\mathbf{U}^\intercal \right)\big),~
	\mathbf{S} = \text{diag}\left(\underbrace{1, \cdots, 1}_{N_{\mathcal{F}}-1}, d \right)
  \end{split}  
  \end{equation}
 \revised{ The optimal rotation matrix ${}^{l}\mathbf{R}_{w}$ and scale factor ${}^{l}s_{w}$ are computed using the SVD of the matrix $\mathbf{H}$:}
  \begin{align}
	  {}^{l}\mathbf{R}_{w} &= \mathbf{U}\mathbf{S}\mathbf{V}^\intercal  \label{eq:R} \\ 
	  {}^{l}s_{w} &= \frac{\sigma_{{}^{w}\mathbf{X}_{c_e}}^2}{\text{tr}(\Sigma \mathbf{S})} \label{eq:s}  
  \end{align} 
 Finally, the translation vector ${}^{l}\mathbf{t}_{w}$ is computed as follows:
  \begin{equation}
	  {}^{l}\mathbf{t}_{w} = {}^{w}\mathbf{\bar{X}}_{c_e} - {}^{l}s_{w}{}^{l}\mathbf{R}_{w}{}^{l}\mathbf{\bar{X}}^{cmc}_{a,o_{t}}
  \end{equation}
 }
 
After that, we use the estimated $\left\{ {}^{l}s_{w},{}^{l}\mathbf{R}_{w},{}^{l}\mathbf{t}_{w} \right \}$ to transform $\left \{ {}^{l}\mathbf{X}^{cmc}_{a,f,o} \right \}$ and $\left \{ \left [{}^{l}\mathbf{R}_{c_i} \middle | {}^{l}\mathbf{X}_{c_i} \right ] \right \}$  from the local coordinate to the world coordinate:
  %\begin{scriptsize}
	\begin{align}
	  {}^{w}\mathbf{X}^{cmc}_{a,f,o} = & {}^{l}s_{w} {}^{l}\mathbf{R}_{w} {}^{l}\mathbf{X}^{cmc}_{a,f,o} + {}^{l}\mathbf{t}_{w} \label{eq:reblanced_matching_1} \\ 
	  {}^{w}\mathbf{R}_{c_i} = &{}^{l}\mathbf{R}_{w} {}^{l}\mathbf{R}_{c_i} \label{eq:reblanced_matching_2} \\
	  {}^{w}\mathbf{X}_{c_i} = &{}^{l}s_{w} {}^{l}\mathbf{R}_{w} {}^{l}\mathbf{X}_{c_i} + {}^{l}\mathbf{t}_{w} \label{eq:reblanced_matching_3}.
	\end{align}  
  %\end{scriptsize}
  We apply RGB-image feature detectors~\cite{detone2018superpoint, sun2021loftr} again to obtain
  2D scene keypoints and their corresponding features for CMC RGB images. Then, we integrate CMC 6-DoF pose priors (\ie, $\left \{ \left [{}^{w}\mathbf{R}_{c_i} \middle | {}^{w}\mathbf{X}_{c_i} \right ] \right \}$)
  to match these 2D keypoints to 3D scene keypoints~\cite{sarlin2019coarse}, which are estimated from our ego-camera processing. All 2D-3D matches, from both scene and mobile keypoints, are merged into a set of 2D-3D pairs represented by
\begin{align}
	\left \{ \left \langle \mathbf{x}_{c_i,f,m}, {}^{w}\mathbf{X}_{m} \right \rangle ~\vert~ \forall c_i \in \mathcal{C}^{cmc} \cup \mathcal{F}, \forall f \in \mathcal{F}, \right. \\
	\nonumber
	\left.
	\forall m \in \mathcal{M}\cup \left\{\mathcal{F}\times\mathcal{O} \right\} \right \}.
\end{align}  
%where $\mathcal{C}^{ego} = \left\{ \left \langle c_{e},f \right \rangle \vert \forall f \in \mathcal{F}\right\} $.

  \noindent{\bf Joint optimization with factor graph}. Within the same coordinate system, Eqs.~\ref{eq:ego_bd} and~\ref{eq:cmc_bd} can be consolidated into a single equation. We use $\varepsilon_{\textrm{c2k}}(\cdot)$ to denote the 2D-3D projection constraints that encompass all camera poses and keypoints. It incorporates a joint optimization of the factors between: CMC poses and mobile-agent keypoints, CMC poses and scene keypoints, ego-camera poses and scene keypoints, and relative constraints of the CMC poses. This is visually represented in Fig.~\ref{fig:factor_graph} and mathematically expressed in the subsequent equation:
  %\begin{scriptsize}
	\begin{equation}\label{eq:c2k}
	\begin{split}
	 \argmin_{ \mathcal{T}, \mathcal{X} } &
	 \sum_{\forall c_i \in \mathcal{C}^{cmc} \cup \mathcal{F}}~
	 \sum_{\forall f \in \mathcal{F}} ~
	 \sum_{\forall m \in \mathcal{M}\cup \{\mathcal{F}\times\mathcal{O}\} }  \varepsilon_{\textrm{c2k}} \left (\mathcal{T}, \mathcal{X}  \right ),
	\end{split}
\end{equation}
where all 3D keypoints and camera poses are merged into the set $\mathcal{X}$ and the set $\mathcal{T}$, respectively. In detail, we represent them by
\begin{equation}
		\begin{split}
  \varepsilon_{\textrm{c2k}} \left ( \mathcal{T}, \mathcal{X}  \right ) =&  v_{c_i,f,m} \left\| \mathbf{x}_{c_i,f,m} - \pi \left (\mathbf{K}_{c_i}\left [{}^{w}\mathbf{R}_{c_i}^{\top} \middle | - {}^{w}\mathbf{R}_{c_i}^{\top} {}^{w}\mathbf{X}_{c_i} \right ] , \right. \right. \\ &\left. \left. {}^{w}\mathbf{X}_{m}\right ) \right\|_2 ,\\
	  \mathcal{T} = &\left \{ \left [{}^{w}\mathbf{R}_{c_i} \middle | {}^{w}\mathbf{X}_{c_i} \right ]\right \} \cup \left \{ \left [{}^{w}\mathbf{R}_{c_e,f} \middle | {}^{w}\mathbf{X}_{c_e,f} \right ] \right \} \\
	  = & \left \{ \left [{}^{w}\mathbf{R}_{c_i} \middle | ~{}^{w}\mathbf{X}_{c_i} \right ] \middle |
	  \forall c_i \in \mathcal{C}^{cmc} \cup \mathcal{F}\right \},  \\
	  \mathcal{X} = & \left \{ {}^{w}\mathbf{X}^{map}_{m} \right \} \cup  \left \{ {}^{w}\mathbf{X}^{cmc}_{a,f,o} \right \} \\
	   = & \left \{ {}^{w}\mathbf{X}_{m} \middle | ~
	  \forall m \in \mathcal{M}\cup \left\{\mathcal{F}\times\mathcal{O} \right\}   \right \} ,
	\end{split}
\end{equation}
  %\end{scriptsize} 

  To complete our factor graph illustrated in Fig.~\ref{fig:factor_graph}, we further define a error function $\varepsilon_{\textrm{a2e}}(\cdot)$. It is designed to model the discrepancy between agent's top keypoints $\left \{ {}^{w}\mathbf{X}^{cmc}_{a,f,o_{t}} \right \}$ and ego-camera positions $\left \{ {}^{w}\mathbf{X}_{c_e,f} \right \}$. Since $\left \{ {}^{w}\mathbf{X}^{cmc}_{a,f,o_{t}} \right \}$ and  $\left \{ {}^{w}\mathbf{X}_{c_e,f}\right \}$ are correlated, we solely apply a kinematic constraint $\varepsilon_{\textrm{kin}}(\cdot)$ to $\left \{ {}^{w}\mathbf{X}^{cmc}_{a,f,o} \right \}$ to jointly model kinematic factors of mobile agent and ego camera. Within the kinematic factors, we hypothesize that the agent's motion vector, and, the vector from the agent's bottom ($o_b$) to top ($o_t$), may exhibit minimal changes between successive frames.
  The formulas are described as follows: 
  %\begin{scriptsize}
	  \begin{align}
		\varepsilon_{\textrm{a2e}}\left ({}^{w}\mathbf{X}_{c_e,f},{}^{w}\mathbf{X}^{cmc}_{a,f,o_{t}} \right ) =& \left \|  {}^{w}\mathbf{X}_{c_e,f}-{}^{w}\mathbf{X}^{cmc}_{a,f,o_{t}} \right \|_2 \label{eq:f_e2e}, \\
	  \varepsilon_{\textrm{kin}} \left ( {}^{w}\mathbf{X}^{cmc}_{a,f-1:f+1} \right ) =& 
	  \begin{aligned}[t] 
		& \left \| \left ( {}^{w}\mathbf{X}^{cmc}_{a,f+1, o_{t}}-{}^{w}\mathbf{X}^{cmc}_{a,f,o_{t}}\right ) \right. \\ &\left.- \left ( {}^{w}\mathbf{X}^{cmc}_{a,f,o_{t}} 
		-{}^{w}\mathbf{X}^{cmc}_{a,f-1, o_{t}} \right ) \right \|_2 \\
		& + \left \| \left ( {}^{w}\mathbf{X}^{cmc}_{a,f+1, o_{t}}-{}^{w}\mathbf{X}^{cmc}_{a,f+1, o_{b}} \right ) \right.\\
		&\left. - \left ({}^{w}\mathbf{X}^{cmc}_{a,f,o_{t}} -{}^{w}\mathbf{X}^{cmc}_{a,f,o_{b}} \right )
		\right \|_2, 
	\end{aligned} \label{eq:f_kin} 
	\end{align}  
  %\end{scriptsize}
  Note that, we apply a simple kinematics constraint in Eq.~\ref{eq:f_kin} to simplify the Jacobian matrix calculation in our custom factors~\cite{dellaert2012factor}. However, to better formulate the kinematics of the mobile agent, we may incorporate the IMUs data~\cite{Lupton2012Visual-Inertial-Aided, campos2021orb} and consider agent's joint angles and limb lengths~\cite{Bioslam2020, bultmann2021real}. We leave these to the future works.

  The factor graph~\cite{dellaert2012factor, dellaert2017factor} is a graphical model that represents the probabilistic relationships between variables, observations, and constraints, facilitating efficient optimization in the scene mapping and camera pose estimation process. Given all observations $\mathcal{Z}$, the posterior probability of all variables $\mathcal{Y}$ is represented as $ \mathbb{P} \left ( \mathcal{Y} \vert \mathcal{Z}  \right )$. In our factor graph optimization, we want to maximize $ \mathbb{P} \left ( \mathcal{Y} \vert \mathcal{Z}  \right )$, which is equal to minimize $- \log \left (\mathbb{P} \left ( \mathcal{Y} \vert \mathcal{Z}  \right ) \right ) $.  This can be further simplified and expressed as
  %\begin{scriptsize}
  \begin{equation}
	\label{eq:factor_graph_all}
	\begin{split}
	  \argmin_{\mathcal{Y}} & \left \{    \sum_{\forall c_i \in \mathcal{C}^{cmc} \cup \mathcal{F}}
	  \sum_{\forall f \in \mathcal{F}}
	  \sum_{\forall m \in \mathcal{M}\cup \{\mathcal{F}\times\mathcal{O}\} }
		 \left \| \varepsilon_{\textrm{c2k}} \right \|^2_{\Sigma_{\textrm{c2k}}}  \right. \\ & \left.+
	  \sum_{\forall f \in \mathcal{F}} \left \| \varepsilon_{\textrm{kin}} \right \|^2_{\Sigma_{\textrm{kin}}} +
	  \sum_{\forall f \in \mathcal{F}} \left \| \varepsilon_{\textrm{a2e}} \right \|^2_{\Sigma_{\textrm{a2e}}} 
	  \right \},\\
	  \textrm{where}~ \mathcal{Y} = & \left \{\mathcal{T}, \mathcal{X} \right \} \\
	  \mathcal{Z} = & \left \{ \mathbf{x}_{c_i,f,m}~\vert~\forall c_i~\in~\mathcal{C}^{cmc}~\cup~\mathcal{F},~~
	  \forall~f~\in~\mathcal{F}, \right. \\
	& \left. \forall~m~\in~\mathcal{M}~\cup~\left \{\mathcal{F}\times\mathcal{O} \right \} \right \}.
  \end{split}  
  \end{equation}
  where  $\Sigma_{\textrm{c2k}}$, $\Sigma_{\textrm{a2e}}$ and $\Sigma_{\textrm{kin}}$ denote the covariance matrices.

  \begin{figure}[th!]
	\centering
	\includegraphics[width=0.9\linewidth]{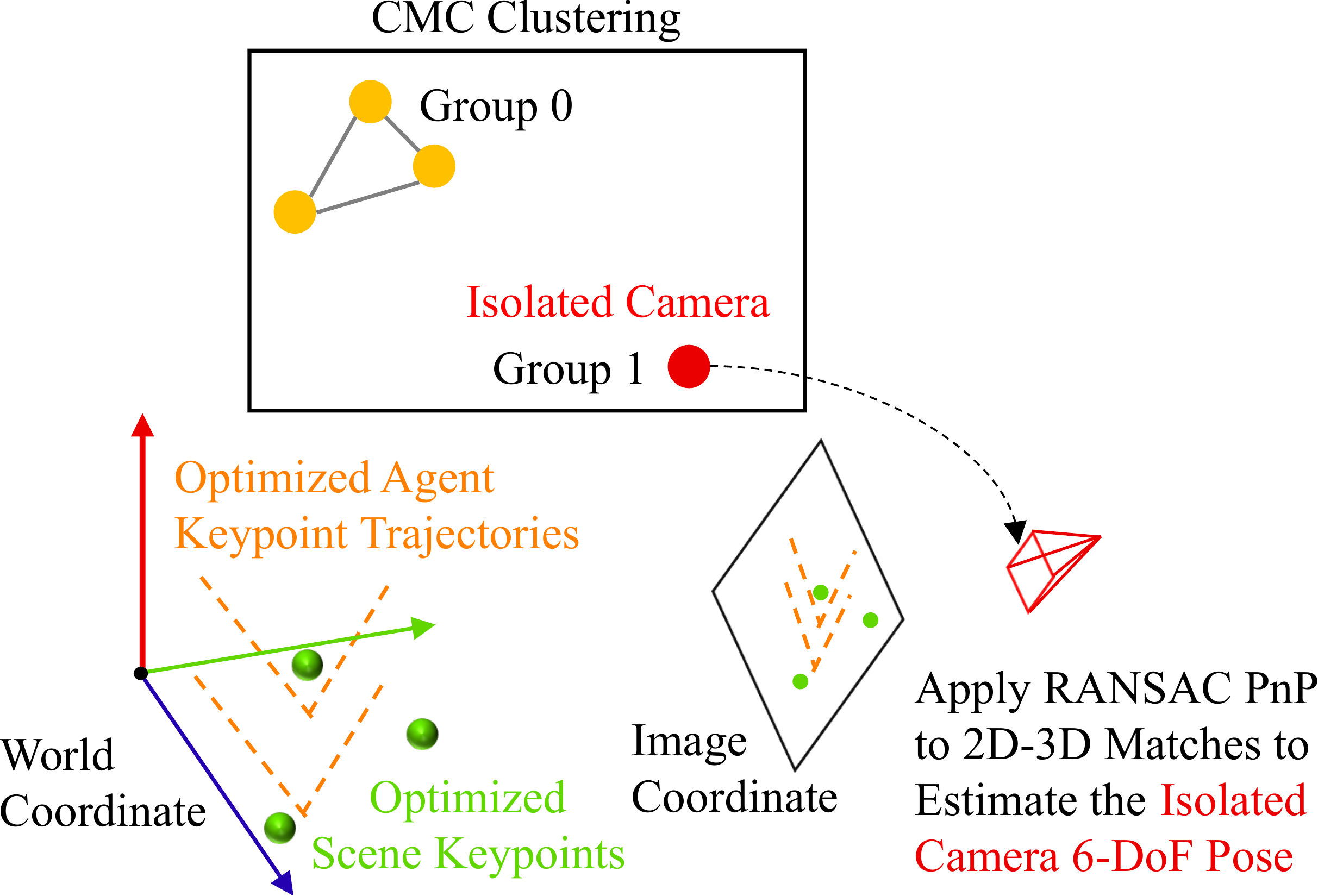}
	\caption{\revised{\textbf{Register an isolated CMC 6-DoF pose.} This process involves the utilization of static scene keypoints and mobile agent keypoints to establish a 2D-3D matches for isolated CMCs. Following this, PnP and RANSAC are performed to estimate the isolated camera 6-DoF pose.}}
	\label{fig:isolated_cam_pose}
  \end{figure}

  Using optimized ego-camera poses and corresponding RGB-D images, we refine the constructed point clouds. We then apply the planar patch detection~\cite{AraujoOliveira_2020a} on refined point clouds to obtain a parameterized scene layout (Fig.~\ref{fig:framework}). Our target outputs, as the parameterized scene layout and CMC 6-DoF poses registered to it, have been achieved.

\subsection{Register 6-DoF Poses for Isolated CMCs}
  \label{sec:register_isolated_CMC}

  \revised{ Although previous processes require at least two overpping-view CMCs to estimate CMC relative 6-DoF poses and pseudo-scale agent keypoint trajectories, there may be isolated CMC that has no overlapping views with others.
Despite this, YOWO can still register such an isolated CMC to the scene layout. Fig.~\ref{fig:isolated_cam_pose} gives an example, of how YOWO registers an isolated CMC to the scene layout. Basically, YOWO processes the isolated CMC after completing the scene mapping and non-isolated CMC registration. It is assumed that the isolated CMC can observe agent mobile keypoints and scene static keypoints, which have been jointly optimized by YOWO in the preceding steps. By referring to the timestamps of constructed 3D mobile keypoints, we can form the 2D-3D matches for observed mobile keypoints. Employing RANSAC and PnP~\cite{fischler1981random, kneip2011novel, zheng2013revisiting}, we initialize the 6-DoF pose of the isolated CMC. Based the initialized 6-DoF pose prior and visual features, we further obtain 2D-3D matches for observed scene static keypoints. We then incorporate the 2D-3D matches of both scene static keypoints and mobile keypoints to refine the isolated CMC 6-DoF pose.
}
  
  %%%%%%%%%%%%%%%%%%%%%%%%%%%%%%%%%%%%%%%%%%%%%%%%%%%%%%%%%%%%%%%%%%%%%%%%%%%%%%%%
  
  \section{Experiments}

  \begin{figure*}[th!]
	\centering
	%\captionsetup{font=scriptsize}
	\includegraphics[width=\textwidth]{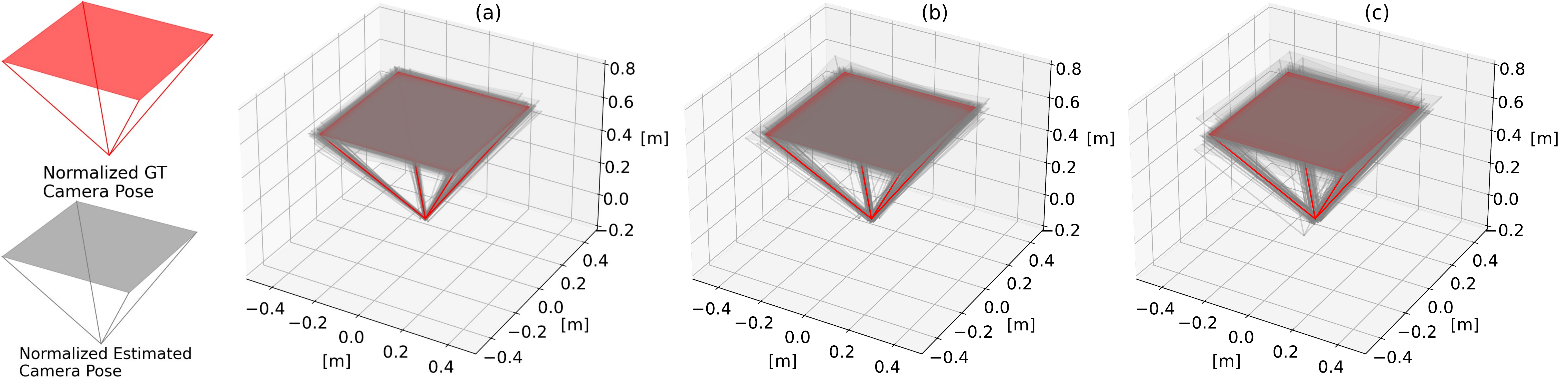}
	\captionof{figure}{\textbf{Visualization of bias between estimated CMC 6-DoF poses and GTs, in a normalized space with all GTs aligned to one pose.} The normalized GT and estimated camera poses are rendered with {\color{red}red} and {\color{gray} gray} colors, respectively. The relative 6-DoF pose estimation bias of methods YOWO, OMECC~\cite{patzold2022online}, and ECCMP~\cite{lee2022extrinsic} are plotted in (a), (b), and (c), respectively.}
	\label{fig:relative_pose_vis}
	%\vspace{-0.5cm}
  \end{figure*}
  
  \subsection{Experimental Setup}
  
  \noindent{\bf Dataset.} As shown in Tab.\ref{table:data_compare}, we contribute a CMC and ego-camera collaborative capturing dataset for jointly evaluating scene layout mapping and CMC 6-DoF pose registration. Utilizing AI2THOR~\cite{kolve2017ai2, procthor} and Gym-UnrealCV~\cite{gymunrealcv2017, qiu2017unrealcv}, we generate indoor scenes that encompass the house, garage, school, etc. An agent is then navigated within these scenes for collaborative capture. Taking  advantage of simulation, we introduce variations in texture, illumination, object presence, and scale within these indoor scenes. This ensures a comprehensive evaluation of our proposed method under a wide array of conditions.

  \noindent{\bf Metrics.}
  We utilize standard metrics commonly used in visual localization~\cite{sattler2018benchmarking, sarlin2019coarse, do2022learning, humenberger2022investigating} to evaluate CMC 6-DoF poses. The valuation metrics are represented by
  %\begin{scriptsize}
	\begin{align}\label{eq:metrics}
	  \varepsilon_{\textrm{pos}} &= \Vert {\mathbf{X}_{c_i} - \mathbf{X}_{c_i}^{\textrm{gt}}} \Vert _2 \\ 
	  \varepsilon_{\textrm{rot}} &= \arccos \left( \frac{\textrm{trace}\left( \mathbf{R}_{c_i}^{\top} \cdot \mathbf{R}_{c_i}^{\textrm{gt}}\right) - 1}{2} \right)
	\end{align}
  %\end{scriptsize}
  Besides, to evaluate the scene mapping performance, we follow~\cite{endres2012evaluation, sturm2012benchmark, zhu2022nice} to
  compare the estimated ego-camera trajectory to GT using ATE RMSE, and follow~\cite{stekovic2021montefloor, yue2023connecting, tabata2023shape}
  to evaluate the 2D top-view scene layout with polygon IoU.

  \subsection{Comparison with State-of-the-Art (SOTA) Methods}
  
  \noindent{\bf YOWO vs. SOTA CMC relative 6-DoF pose estimation.} 
  ECCMP~\cite{lee2022extrinsic} and OMECC~\cite{patzold2022online} are representative works that utilize mobile keypoints for CMC relative 6-DoF pose estimation. Although their estimated CMC relative 6-DoF poses are not correlated with the scene layout and are isolated within each room, we follow their evaluation protocol to align CMC relative 6-DoF poses to the GT CMC layout for the evaluation. As shown in Tab.\ref{table:relative_pose_eval} and Fig.~\ref{fig:relative_pose_vis}, all methods yield promising results by leveraging the mobile keypoints. Our YOWO method, however, outperforms all others in every metric, albeit the improvement is marginal. The rationale here is that mobile keypoints provide robust cues for cross-camera matching, which plays a vital role in CMC relative 6-DoF pose estimation. Besides, due to the visual ambiguity (Fig.~\ref{fig:visual_ambiguity}), even though YOWO further incorporates static scene keypoints into its joint optimization, the improvement is marginal compared to ECCMP~\cite{lee2022extrinsic} and OMECC~\cite{patzold2022online} that solely rely on mobile keypoints. Nonetheless, it is important to note that, the CMC relative 6-DoF poses possess a pseudo scale and are not automatically aligned with the scene layout, hindering their practical application. Hence, it is recommended to move to the CMC 6-DoF pose registration, which requires automatically aligning CMC 6-DoF poses to the scene layout.

  \begin{table}[th!]
	\begin{center}
	  \setlength{\tabcolsep}{5.2pt}
	  %\scriptsize
	  %\captionsetup{font=scriptsize}
	  \caption{\textbf{Comparison with SOTAs on CMC relative 6-DoF pose estimation.} The relative 6-DoF poses are aligned to the GT CMC layout by using the KU algorithm~\cite{kabsch1976solution, umeyama1991least}. The identical mobile keypoints are used for all methods. The best and the second-best results in each metric are highlighted in \textbf{bold} and \underline{underlined}, respectively. }
	  \label{table:relative_pose_eval}
	  \begin{tabular}{lcccccc}
		\multirow{ 2}{*}{Method}   & \multicolumn{3}{c}{$\varepsilon_{\textrm{pos}}$ [m] $\downarrow$} & \multicolumn{3}{c}{$\varepsilon_{\textrm{rot}}$ [deg] $\downarrow$}  \\ 
		\cmidrule(lr){2-4} \cmidrule(lr){5-7} 
		& Avg.  & Min.  & Max. & Avg.  & Min.  & Max.
		\\\toprule
	  ECCMP  & 0.024 & \textbf{0.002}  & 0.105 & 0.338 & \underline{0.097} & 0.811\\  
	  OMECC & \underline{0.015} & \textbf{0.002} & \underline{0.062} &\underline{0.216} & 0.103 & \underline{0.529} \\ 
	  \textbf{YOWO}       & \textbf{0.008} & \textbf{0.002} & \textbf{0.033} & \textbf{0.135} & \textbf{0.092} & \textbf{0.359}\\ 
	  \end{tabular}
	\end{center}
  \end{table}

  \begin{figure*}[th!]
	\begin{minipage}[b]{.48\linewidth}
	  \centering
	  \setlength{\tabcolsep}{1.6pt}
	  %\scriptsize
	  %\captionsetup{font=scriptsize}
	  \captionof{table}{\textbf{Comparison with SOTAs on CMC 6-DoF pose registration.} Since the estimated scene layouts have been aligned to their GTs, the registered CMC 6-DoF poses are directly compared to their GTs. The best results in each metric are highlighted in \textbf{bold}. }
	  \label{table:register_pose_eval_1}
	  \begin{tabular}[b]{lcccccc}
		\multirow{ 2}{*}{Method}   & \multicolumn{3}{c}{$\varepsilon_{\textrm{pos}}$ [m] $\downarrow$} & \multicolumn{3}{c}{$\varepsilon_{\textrm{rot}}$ [deg] $\downarrow$}  \\ 
		\cmidrule(lr){2-4} \cmidrule(lr){5-7} 
		& Avg.  & Min.  & Max. & Avg.  & Min.  & Max.
		\\\toprule
		Hloc & 1.345 & 0.014& 11.132 & 14.958 & 0.367 & 54.231   \\
		Kapture & 1.187 & 0.011& 10.787 & 12.635 & 0.236 & 57.958\\
		\textbf{YOWO}       & \textbf{0.131}  & \textbf{0.007} & \textbf{0.348} & \textbf{0.205} & \textbf{0.026} & \textbf{0.762} \\ 
	  \end{tabular} 
	\end{minipage}\hfill
	\begin{minipage}[b]{.48\linewidth}
	  \centering
	  \setlength{\tabcolsep}{1.6pt}
	  %\scriptsize
	  %\captionsetup{font=scriptsize}
	  \captionof{table}{\textbf{Evaluation of SOTAs using YOWO's CMC 6-DoF pose priors to assist their 2D-3D matches.} By using YOWO CMC 6-DoF pose priors and the 3D scene model, Hloc and Kapture can circumvent the vulnerable global search and effectively construct 2D-3D matches.}
	  \label{table:register_pose_eval_2}
	  \begin{tabular}[b]{lcccccc}
		  \multirow{ 2}{*}{Method}   & \multicolumn{3}{c}{$\varepsilon_{\textrm{pos}}$ [m] $\downarrow$} & \multicolumn{3}{c}{$\varepsilon_{\textrm{rot}}$ [deg] $\downarrow$}  \\ 
		  \cmidrule(lr){2-4} \cmidrule(lr){5-7} 
		  & Avg.  & Min.  & Max. & Avg.  & Min.  & Max.
		  \\\toprule
		Hloc &  0.259 & 0.010 & 0.678 & 0.711 & 0.149 & 1.247\\
		Kapture & 0.213 & 0.008 & 0.634 & 0.585 & 0.122 & 1.321\\
		\\
	  \end{tabular} 
	\end{minipage}
  \end{figure*}

  \begin{figure*}[ht!]
	\centering
	\includegraphics[width=\textwidth]{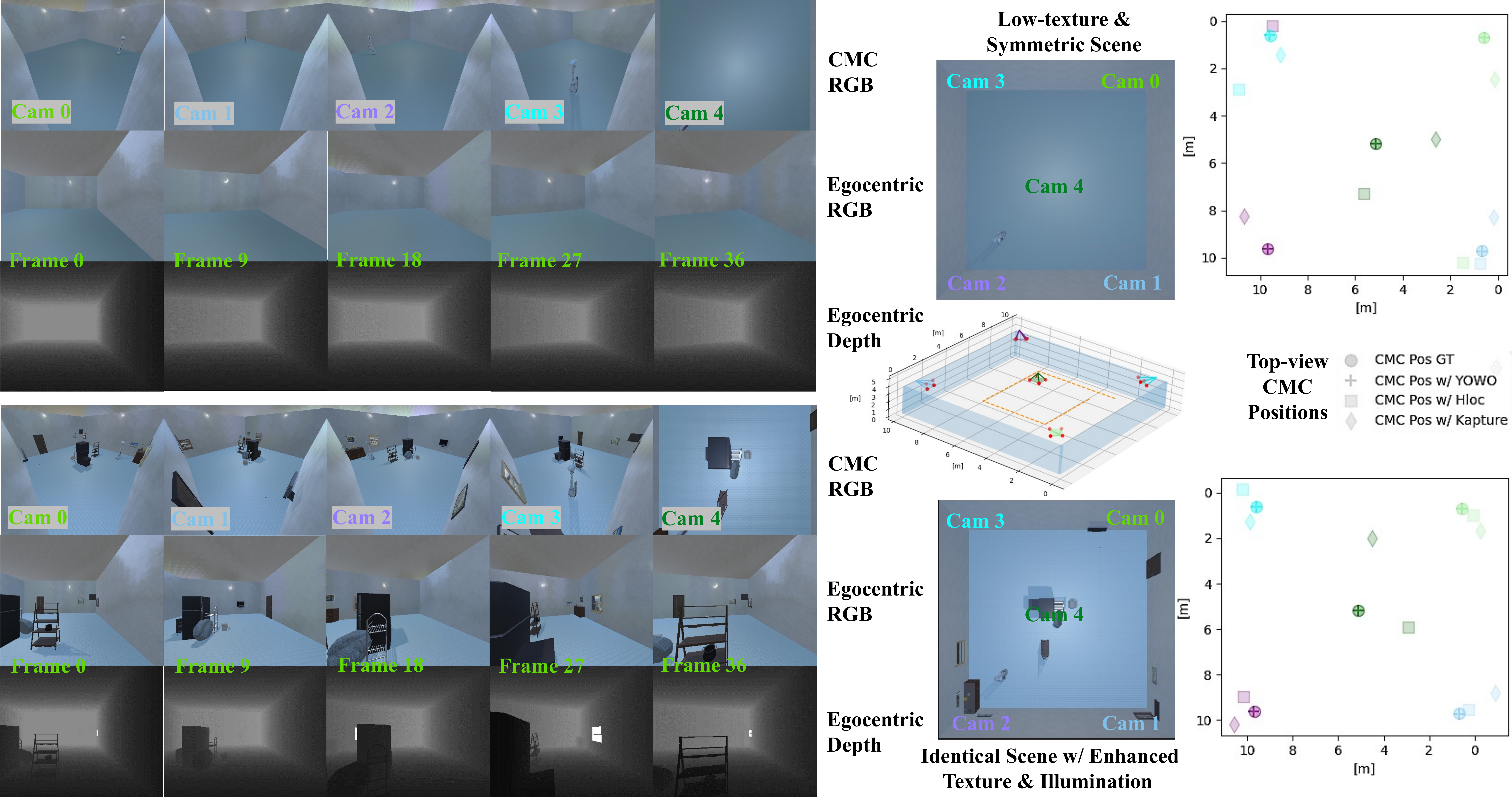}
	\caption{\textbf{Comparison between YOWO, Hloc~\cite{sarlin2019coarse}, and Kapture~\cite{humenberger2020robust, humenberger2022investigating} for CMC 6-DoF pose registration.} Using the identical room, we specifically focus on \revised{analyzing} the effect of various illumination and texture, and further infer the robustness of each model.}
	\label{fig:ai2thor_scene_45}
	%\vspace{-0.5cm}
  \end{figure*}

  \begin{table}[th!]
	\begin{center}
	  \setlength{\tabcolsep}{.5pt}
	  %\scriptsize
	  %\captionsetup{font=scriptsize}
	  \caption{\textbf{Comparison with SOTAs on scene mapping.} The best and the second-best results in each metric are highlighted in \textbf{bold} and \underline{underlined}, respectively.}
	  \label{table:scene_map}
	  \begin{tabular}{lcc}
		& \begin{tabular}[c]{@{}l@{}}Ego-camera\\ ATE RMSE [m]\end{tabular} $\downarrow$ & Layout IoU $\uparrow$ \\ \toprule
	  BAD-SLAM & 0.286  & 0.848 \\ 
	  NICE-SLAM& \underline{0.252}   & \underline{0.873} \\ 
	  YOWO & \textbf{0.122}   &  \textbf{0.927} \\ %\hline
	  \end{tabular}
	\end{center}
  \end{table} 

  \noindent{\bf YOWO vs. SOTA scene mapping.} We compare YOWO with a traditional RGB-D SLAM method---BAD-SLAM~\cite{Schops_2019_CVPR}, and a more recent solution---NICE-SLAM~\cite{zhu2022nice}, which utilizes neural implicit representation. For all methods, we apply a planar detector~\cite{AraujoOliveira_2020a} to parameterize obtained point clouds to scene layouts. The results are presented in Tab.~\ref{table:scene_map}. Unlike other visual localization datasets~\cite{sturm2012benchmark, dai2017scannet}, our dataset features a fast-moving agent that may not form loop closures, which often leads to drift errors when RGB-D SLAM methods are employed. While BAD-SLAM performs on par with NICE-SLAM, YOWO outperforms both, benefiting from the incorporation of CMC observations to enhance the ego-camera scene mapping. Although we evaluate scene mapping performance, YOWO is not tied to a specific SLAM method and can adapt to incorporate more advanced SLAM methods in the future.

  %However, optimizing the agent path and utilizing distinct visual scene features could enhance the performance of NICE-SLAM and BAD-SLAM.
  
  %Thus, we demonstrate that incorporating CMC observations enhances the ego-camera scene mapping. 

  \begin{table*}[th!]
	\begin{center}
	  \setlength{\tabcolsep}{1.5pt}
	  %\scriptsize
	  %\captionsetup{font=scriptsize}
	  \caption{\textbf{Ablation studies on YOWO.} See Fig.~\ref{fig:framework}, three processes of YOWO include: (i) ego-camera processing, (ii) CMC processing, and (iii) collaborative processing. JO denotes the joint optimization with a factor graph using Eq.~\ref{eq:factor_graph_all}.}
	  \label{table:ablation}
	  \begin{tabular}{lcccc}
		\multirow{ 2}{*}{}  & \multicolumn{2}{c}{CMC registration} & \multicolumn{2}{c}{Scene mapping } \\   
		\cmidrule(lr){2-3} \cmidrule(lr){4-5} 
	  & Avg. $\varepsilon_{\textrm{pos}}$ [m] $\downarrow$ & Avg. $\varepsilon_{\textrm{rot}}$ [deg] $\downarrow$  & Ego-camera ATE RMSE [m] $\downarrow$ & Layout IoU $\uparrow$ \\ \toprule
	  (i)    & - & - & 0.165 & 0.915 \\ 
	  (i)(ii)(iii) w/o JO & 0.173  & 0.318  & 0.165 & 0.915 \\ 
	  (i)(ii)(iii) w/ JO & 0.131 ({\color{violet}{-0.042}})   & 0.205 ({\color{violet}{-0.113}})   & 0.122 ({\color{violet}{-0.043}})  & 0.927 ({\color{violet}{+0.012}}) \\ %\hline
	  \end{tabular}
	\end{center}
  \end{table*}

  \begin{figure*}[th!]
	\centering
	%\captionsetup{font=scriptsize}
	\includegraphics[width=\textwidth]{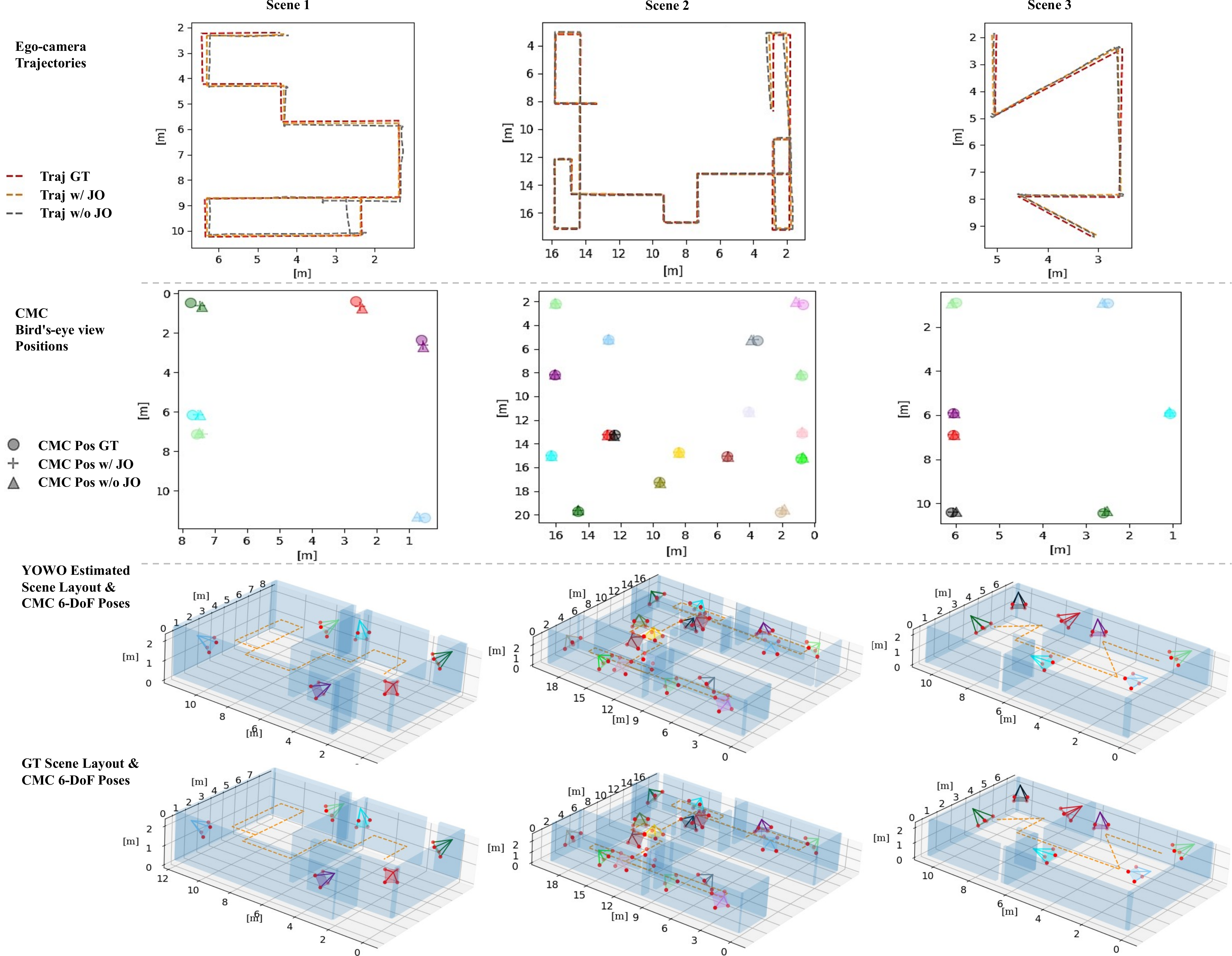}
	\captionof{figure}{\textbf{Visualization of ablation study results.} Three indoor scenes are illustrated as examples. In each scene, we plot ego-camera trajectories and CMC positions for the GT, the estimations w/ JO, and the estimations w/o JO, respectively. }
	\label{fig:vis_ablation}
	%\vspace{-0.5cm}
  \end{figure*}

  \noindent{\bf YOWO vs. SOTA CMC 6-DoF pose registration.} We compare YOWO with SOTA visual localization methods: Hloc~\cite{sarlin2019coarse} and Kapture~\cite{humenberger2020robust, humenberger2022investigating}.
  For a fair comparison, all methods utilize our estimated ego-camera poses to construct the identical real-scale 3D scenes, as opposed to employing pseudo-scale ego-camera poses estimated by COLMAP~\cite{schonberger2016structure}. Notably, the scene reconstruction bias is inherently transferred to the CMC 6-DoF pose registration, aligning more closely with real-world application scenarios. 
  
  As shown in Tab.~\ref{table:register_pose_eval_1},
  using the identical 3D scene model, YOWO delivers a performance leap over Hloc and Kapture on CMC 6-DoF pose registration. Both Hloc and Kapture have their maximum $\varepsilon_{\textrm{pos}}$ and $\varepsilon_{\textrm{rot}}$ exceeding $10$ meters and $50$ degrees, respectively. In contrast, YOWO maintains these errors within $0.4$ meters and $0.8$ degrees, respectively. 
  For Hloc and Kapture, the quality of CMC 6-DoF pose registration heavily relies on global search (\ie, image retrieval). However, to make their global search work properly, the query CMC viewpoint should be similar to one of the ego-camera reference viewpoints, which is challenging as the CMC positions may not be accessible to the ego camera (Fig.~\ref{fig:visual_ambiguity}). Therefore, remarkable errors are reported in Tab.~\ref{table:register_pose_eval_1}. Moreover, those errors could potentially increase when applied to larger indoor scenes.

  We also conduct additional experiments, as detailed in Tab.~\ref{table:register_pose_eval_2}, to investigate the feasibility of applying YOWO's CMC 6-DoF pose priors to assist 2D-3D matches for Hloc and Kapture. The results indicate that the performance of Hloc and Kapture improves considerably, nearly matching the level of YOWO. This reveals that YOWO's CMC 6-DoF pose priors can bestow enhanced robustness upon existing visual localization methods.
  However, unlike YOWO, Hloc and Kapture are unable to include mobile keypoints to perform a joint optimization with scene static keypoints. As a result, even with YOWO's CMC 6-DoF pose priors, Hloc and Kapture still exhibit slightly weaker performance than YOWO.

  In the end, we include examples in Fig.~\ref{fig:ai2thor_scene_45} to explore how the scene appearance affect the robustness of each model. In an indoor scene with a highly symmetrical structure and ambiguous texture, our YOWO can obtain accurate 6-DoF CMC poses by leveraging mobile keypoints. On the other hand, such a challenging scene can easily cause conventional visual localization to fail, as it is difficult to correctly determine the correspondence between ego-camera and CMC captures by solely relying on the static scene features. After introducing new objects and adjusting the illumination, the static scene features became more distinctive. Consequently, the performance of traditional visual localization methods has seen significant improvement. However, for conventional visual localization to work properly, the viewpoint of the observed CMC must be similar to one of the ego-camera's reference viewpoints, which is difficult because the CMC's positions may not be accessible to the ego-camera. As a result, we can still observe notable biases in some of their results. In a nutshell, YOWO can generate high-performance results in both scenes, which reveals the robustness of YOWO.

  \begin{figure*}[t!]
	\centering
	\includegraphics[width=\textwidth]{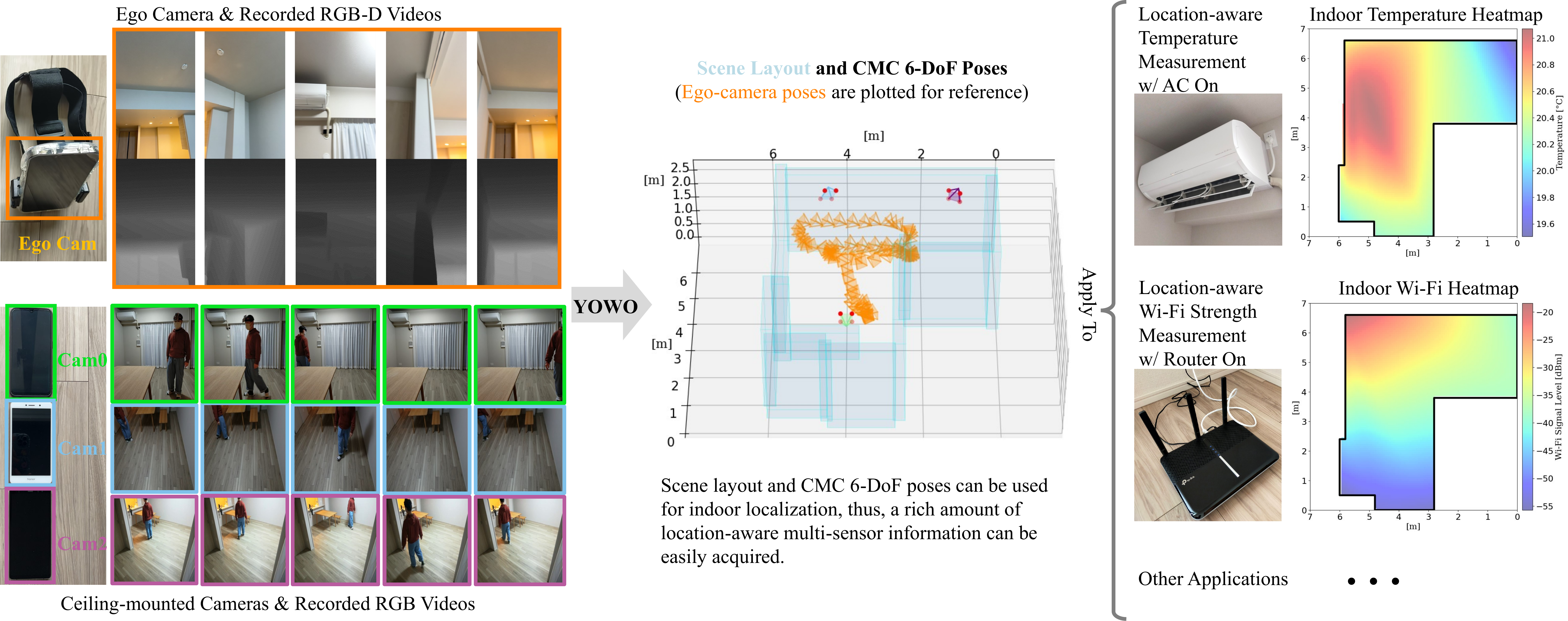}
  %\captionsetup{font=scriptsize}
	\caption{\revised{\textbf{Downstream applications of YOWO in real-life environments.} We capture data in a real indoor scene with four synchronized phone cameras: one for the ego camera and three for CMCs. We then apply YOWO to obtain the scene layout and register CMCs to the scene, which facilitate the collection of position-aware multi-sensor data for downstream applications.}}
  \label{fig:downstream_app}
  \end{figure*}

  \subsection{Ablation Studies}
  
  We conduct ablation studies to validate the effectiveness of joint optimization in YOWO. 
  The scene mapping performance is initially assessed using only ego-camera processing. Subsequently, the CMC 6-DoF registration is evaluated by directly aligning CMC processing results with the raw scene layout. Finally, the contribution of jointly optimizing scene mapping and CMC 6-DoF registration using a factor graph is evaluated.

  While we have shown that unifying scene mapping and CMC 6-DoF pose registration in a single framework overcomes existing challenges (Tabs.~\ref{table:relative_pose_eval},\ref{table:scene_map},\ref{table:register_pose_eval_1},\ref{table:register_pose_eval_2}), the results presented in Tab.~\ref{table:ablation} and Fig.~\ref{fig:vis_ablation} further demonstrate that customizing a factor graph for joint optimization can narrow the gap between those two tasks and enhance the overall performance. Through jointly optimizing scene mapping and CMC 6-DoF pose registration, both agent trajectories and CMC positions more closely align with the ground truths in listed examples of Fig.~\ref{fig:vis_ablation}. Consequently, the estimated scene layouts are also presented appropriate structures as illustrated. To this end, the overall effectiveness and efficiency of our YOWO has been verified.

\subsection{Proof-of-concept Downstream Experiments}
\label{sec:downstream_app}
  
  \revised{The results produced by YOWO, including scene layout and registered CMCs, have the potential to advance a plethora of downstream studies. The direct application of YOWO delivers the capability to automatically model scenes and CMCs, thereby enhancing the efficiency of previous research that employ CMCs for indoor visual observation. It can also assist with indoor scene mapping tasks. For intermediate applications, calibrated CMCs can be leveraged to estimate the 3D object pose under the multi-camera triangulation scheme~\cite{hartley2003multiple}. This allows for efficient collection of position-aware multi-sensor data in a challenging indoor scene. Consequently,  we are capable of supplying 3D body pose labels, encompassing pose estimation, trajectory estimation, and action recognition~\cite{3d_Capture_2017, dai2021indoor, ar_2024}. }
  
  \revised{Additionly, we provide proof-of-concept experiments in Fig.~\ref{fig:downstream_app}. These experiments extend the downstream applications of YOWO beyond the realm of computer vision domain, demonstrating its applicability to other forms of sensing such as Wi-Fi signals and temperature measurements. Note that, some potential downstream applications of YOWO might involve using CMCs to capture images of individuals in public spaces. We advocate for careful ethical considerations in such scenarios.}

  %%%%%%%%%%%%%%%%%%%%%%%%%%%%%%%%%%%%%%%%%%%%%%%%%%%%%%%%%%%%%%%%%%%%%%%%%%%%%%%%

  %%%%%%%%%%%%%%%%%%%%%%%%%%%%%%%%%%%%%%%%%%%%%%%%%%%%%%%%%%%%%%%%%%%%%%%%%%%%%%%%
  
  \section{Discussion}
  
  \noindent{\bf Limitations and applicable scopes.} (i) YOWO works on the assumption of a single agent in the scene, which limits its applicability in indoor scenes with multiple agent-like entities. Thus, when the mobile agent is a person, YOWO is best utilized in people-free settings, such as a supermarket after business hours. While this may seem limiting, it offers two key benefits. Firstly, it avoids the potentially offensive use of an ego camera to capture unknown individuals; secondly, it mitigates the risk of cross-camera association errors.
  (ii) In YOWO, the performance of the ego-camera SLAM and the CMC relative pose estimation are affected by the agent's path. It determines the scene captures for the ego-camera SLAM, and the spatiotemporal distribution of mobile keypoints for the CMC processing. By examining YOWO results in Tabs.~\ref{table:relative_pose_eval},~\ref{table:scene_map}, and~\ref{table:register_pose_eval_1}, we deduce that the ego-camera SLAM primarily influences the CMC registration bias. Therefore, additional efforts are required to plan the agent's path. The objective is to increase the overlapping score as outlined in Eqs.~\ref{eq:grid} and \ref{eq:overlap} for the CMC processing, and to promote loop closures for the ego-camera SLAM. 
  \revised{Since the outputs of YOWO primarily serve downstream applications and YOWO accommodates offline processing, it is feasible to adopt an iterative approach for optimizing the path planning. Specifically, we can execute YOWO multiple times, employing a feedback loop where initial results from the 3D scene modeling and CMC observations are used to guide subsequent adjustments to the capture path.}
 
  %In this study, we plan the agent's path heuristically, but for optimal use of YOWO, automation and optimization of path planning should be considered. 

  \noindent{\bf Conclusion and potential impacts.} We introduce YOWO, the first-of-its-kind framework that can jointly map an indoor scene and register CMCs to the scene layout. The CMCs could be either Closed-Circuit Television (CCTV) cameras or simply smartphone cameras, providing flexibility for use in various indoor environments. Additionally, we propose an inaugural dataset to reveal the synergy between ceiling-mounted cameras and the ego camera, establishing the first benchmark for collaborative scene mapping and CMC 6-DoF registration. Leveraging mobile keypoints of a mobile agent, YOWO surmounts the visual ambiguity challenge in visual localization, and reduces the uncertainty inherent in indoor SLAM. Hence, YOWO underpins an efficient and robust solution for indoor mapping and 3D visual positioning.

\ifCLASSOPTIONcaptionsoff
  \newpage
\fi

\bibliographystyle{IEEEtran}
\bibliography{ref}

\end{document}